\documentclass[preprint,3p,times,twocolumn,authoryear]{elsarticle}

\usepackage{times}
\usepackage{soul}
\usepackage{url}
\usepackage[hidelinks]{hyperref}
\usepackage[utf8]{inputenc}
\usepackage[small]{caption}
\usepackage{graphicx}
\usepackage{xcolor}
\usepackage{color}
\usepackage{amsmath,amssymb,amsfonts}
\usepackage{booktabs}
\usepackage{algorithm}
\usepackage{algorithmic}
\usepackage[switch]{lineno}
\usepackage{algorithmic}
\usepackage{algorithm}
\usepackage{subcaption}
\usepackage[algo2e]{algorithm2e}
\usepackage{multirow}

\newtheorem{theorem}{Theorem}
\newtheorem{definition}{Definition}
\newtheorem{remark}{Remark}

\makeatletter
\renewcommand*{\@opargbegintheorem}[3]{\trivlist
      \item[\hskip \labelsep{\bfseries #1\ #2}] \textbf{(#3)}\ \itshape}
\makeatother

\newcommand{\bx}{\mbox{\boldmath $x$}}

\newcommand{\bW}{\mbox{\boldmath $W$}}

\newcommand{\bX}{\mbox{\boldmath $X$}}
\newcommand{\bY}{\mbox{\boldmath $Y$}}

\newcommand{\NA}{---}

\newcommand{\Real}{\mathbb R}

\newcommand{\be}{\begin{eqnarray}}
\newcommand{\ee}{\end{eqnarray}}

\hypersetup{
colorlinks=true,
linkcolor=red,
urlcolor=purple,
citecolor=blue
}

\begin{document}

\begin{frontmatter}



\title{Reduced storage direct tensor ring decomposition for convolutional neural networks compression}


\author[label1]{Mateusz Gabor\corref{cor1}}
\ead{mateusz.gabor@pwr.edu.pl}
\author[label1]{Rafal Zdunek}

\affiliation[label1]{organization={Faculty of Electronics, Photonics, and Microsystems, Wroclaw University of Science and Technology},
            addressline={Wybrzeze Wyspianskiego 27}, 
            city={Wroclaw},
            postcode={50-370}, 
            country={Poland}}
\cortext[cor1]{Corresponding author}

\begin{abstract}
Convolutional neural networks (CNNs) are among the most widely used machine learning models for computer vision tasks, such as image classification. To improve the efficiency of CNNs, many CNNs compressing approaches have been developed. Low-rank methods approximate the original convolutional kernel with a sequence of smaller convolutional kernels, which leads to reduced storage and time complexities. In this study, we propose a novel low-rank CNNs compression method that is based on reduced storage direct tensor ring decomposition (RSDTR). The proposed method offers a higher circular mode permutation flexibility, and it is characterized by large parameter and FLOPS compression rates, while preserving a good classification accuracy of the compressed network. The experiments, performed on the CIFAR-10 and ImageNet datasets, clearly demonstrate the efficiency of RSDTR in comparison to other state-of-the-art CNNs compression approaches.
\end{abstract}



\end{frontmatter}


\section{Introduction}
\label{sec:introduction}
Convolutional neural networks (CNNs) can be considered as one of the dominant classes of deep learning methods, with powerful applications in computer vision, including image classification (\cite{krizhevsky2012imagenet,he2016deep}), image segmentation (\cite{long2015fully,ronneberger2015u}) or object detection (\cite{newell2016stacked}). 

With the increasing efficiency of CNNs, the size and the number of layers increase as well, which implies a larger number of parameters to save (neural network weights) and a larger number of floating-point operations per second (FLOPS) to process the input image. This topic is very important for mobile and edge devices. Due to their small size, they have limited storage capacity and are less efficient than standard computers. In addition, image processing has to be done in real time on a device. This is crucial, for example, in autonomous cars, where the camera has to detect obstacles very quickly. 

This problem can be addressed in two ways. The first approach is to use more efficient hardware that can store more data and process inputs to CNNs quickly. The other is the software approach that takes advantage of intrinsic over-parameterization of neural networks. As a result, there exists the possibility of compressing them into smaller and more efficient variants. 

There are many CNNs compression methods that fall into the primary areas of pruning, quantization, and low-rank approximations. In this study, we investigate the third approach, in which CNN weights are approximated with matrix/tensor decompositions.
We propose a new approach to this methodology, which is based on reduced storage direct tensor ring decomposition (RSDTR), where the tensor ring (TR) representation with the smallest storage cost is selected among all possible variants at a given decomposition accuracy. Previous studies (\cite{wang2018wide,li2021heuristic,cheng2020novel}) focused on the tensorized version of TR for CNN compression without using the TR decomposition algorithm (\cite{zhao2016tensor}). These approaches significantly reduced the number of parameters of neural networks, but at the cost of a higher number of floating-point operations and a worse quality of the network. In this study, we propose a new simple and efficient approach for CNN compression, which, using the special properties of TR decomposition, allows us to find the TR representation with the smallest number of parameters. For the proposed approach, the compression of parameters goes along with the compression of FLOPS, and the drop in classification accuracy is lower than in previously proposed TR methods. Furthermore, because of the use of a decomposition algorithm, compressed networks can be fine-tuned from decomposed factors instead of training them from scratch, as was done in the previous TR approaches. The proposed method was evaluated using three medium-scale and two large-scale CNNs on two datasets, CIFAR-10 and ImageNet. The results presented in this work show that the proposed approach is very effective in compressing CNNs and is very competitive with other state-of-the-art CNN compression approaches. 

\section{Related works}
\label{sec:related_works}
The most popular CNNs compression methods are based on pruning. The main idea of pruning is to remove redundant connections between layers. In this way, the compressed representation of the neural network is obtained. The pruning was started from approaches such as optimal brain damage (\cite{lecun1989optimal}), in which unimportant connections between layers are identified based on second-order derivatives. Nowadays, there are many other pruning approaches, such as filter-based pruning (\cite{lin2020hrank,li2016pruning,li2019compressing,zhen2023rasp,qian2023hierarchical}), which aim to obtain sparse filters in CNNs, where in most cases the compressed network after pruning is fine-tuned. Quantization is another approach in which CNN weights are represented in a lower precision format (\cite{chen2015compressing}). 

Slightly less popular but not less important compression methods use low-rank approximations that are based on matrix/tensor factorizations (\cite{lebedev2014speeding,kim2015compression}). They can be divided into {\it direct} and {\it tensorization} methods. The former uses the decomposed factors as initial new weights, and hence they can be fine-tuned. The latter is based on hand-designed tensorized neural networks where the weights have tensor model structures, and these methods are mostly trained from scratch. 

Direct methods started from the work of \cite{lebedev2014speeding}, who used the CANDECOMP/PARAFAC (CP) decomposition to compress the convolutional kernel. However, the authors limited their study to compressing only one layer of the AlexNet network. Using CP decomposition, it is difficult to compress all convolutional layers at once and then fine-tune the whole compressed network. Therefore, instead of compressing all layers at once, \cite{astrid2017cp} compressed all convolutional layers in CNNs by alternately compressing and fine-tuning each layer separately. \cite{kim2015compression} used the Tucker decomposition for the first time to compress all convolutional layers and fine-tune them at once. In this approach, the weight tensors are decomposed only with respect to two modes that are related to the input and output channels (Tucker-2) because the other two modes, related to filter size, have low dimensions.  However, the Tucker decomposition is characterized by a relatively large core tensor, which means that more parameters must be stored. To improve the storage efficiency of the Tucker decomposition, \cite{gabor2023compressing} proposed to compress the convolutional kernel using the hierarchical Tucker-2 (HT-2) decomposition, in which the 4-th-order core tensor is further decomposed into smaller 3-rd-order core tensors. Convolutional weight tensors can also be decomposed with the TT model, and this approach was developed by \cite{gabor2022convolutional}. In the literature, we can find many combinations and modifications of direct tensor decomposition methods. One of such approaches is automatic multi-stage compression (\cite{gusak2019automated}), where CNN compression is alternately performed with fine-tuning. Another approach is the Tucker-2-CP method proposed by \cite{phan2020stable}, where the core tensor of the Tucker-2 method is further decomposed according to the CP model. The more general approach is nested decomposition (\cite{zdunek2022nested}), which can be applied to any tensor decomposition-based approach.

On the other hand, the \textit{tensorization} methods do not use any decomposition algorithm to compress CNNs, and the pre-trained layers are designed manually. The first tensorization approach to represent CNNs was proposed by \cite{garipov2016ultimate} who represent convolutional kernels in a higher-order tensor train (TT) format. Another approach is named block-term neural networks (\cite{ye2020block}), where the authors tensorized the weights of convolutional and recurrent neural networks in block-term decomposition format. The TR format was first proposed by \cite{wang2018wide}, where ResNets were compressed using a self-designed TR format with fixed and equal ranks (the rank is the same for all layers in CNN). \cite{cheng2020novel} investigated reinforcement learning to find TR ranks. However, the founded ranks are the same within one layer and differ only between layers. The heuristic rank selection for TR compressed networks was proposed by \cite{li2021heuristic}, where TR ranks were founded using a genetic algorithm. In this work, different ranks were used within one layer, but only for very small CNNs. For larger CNNs, such as ResNets, the ranks were selected only per layer, not for each mode.

\section{Background and Preliminaries}
\label{sec:background}
{\it Notation:} Multi-way arrays, matrices, vectors, and scalars are denoted by calligraphic uppercase letters (e.g., $\mathcal{X}$), boldface uppercase letters (e.g., $\bX$),  lowercase boldface letters (e.g., $\bx$), and unbolded letters (e.g., $x$), respectively. Multi-way arrays, known also as multidimensional arrays, will be referred to as tensors (\cite{Kolda08}).  

\begin{definition}[Tensor contraction]
\label{def_1}
Tensor contraction is a generalized multiplication operation over two arrays, where matrix--matrix and tensor--matrix products are of its special cases. Let $\mathcal{X} \in \Real^{I_1 \times I_2 \times \ldots \times I_N}$ and
$\mathcal{Y} \in \Real^{J_1 \times J_2 \times \ldots \times J_M}$, where $I_n=J_m$.
The mode-$\binom{m}{n}$ contraction of tensors $\mathcal{X}$ and
$\mathcal{Y}$ is defined as $\mathcal{Z} = \mathcal{X} \times_n^m \mathcal{Y}$, where $\mathcal{Z} \in \Real^{I_1 \times \ldots \times I_{n-1} \times I_{n+1} \times
\ldots \times I_N \times J_1 \times \ldots \times J_{m-1} \times J_{m+1} \times \ldots \times J_M}$. 
For the matrices: $\bX \times_2^1 \bY= \bX \bY$.
\end{definition}

\begin{definition}[Multi-mode tensor contraction]
\label{def_2}
 It is an extended version of the tensor contraction, where multiple modes can be contracted. For any $\mathcal{X} \in \Real^{I_1 \times \ldots \times I_N}$ and
$\mathcal{Y} \in \Real^{J_1 \times \ldots \times J_M}$, it can be defined as: $\mathcal{Z} = \mathcal{X} \times_{n_1, \ldots, n_L}^{m_1, \ldots, m_L} \mathcal{Y}$, where $\forall l: I_{n_l} = I_{m_l}$, and $l = 1, \ldots, L$.
\end{definition}

\begin{definition}[Circular shift (\cite{mickelin2020algorithms})]
\label{def_3}
Let $\gamma = [1, N,N-1, \ldots, 2]$ be a generator in cycle notation, $\mathcal{S}_N$ be a set of all circular shifts on $N$ variables, and $\forall k: \tau_k \in \mathcal{S}_N$, $\tau_k = \gamma^{k}$ corresponds to a circular shift to the left by $k$ steps. For $\mathcal{Y} \in \Real^{I_1 \times I_2 \times \ldots \times I_N}$, $\mathcal{Y}^{\tau_k} \in \Real^{I_{k+1} \times \ldots \times I_N \times I_1 \times \ldots \times I_k}$. 
\end{definition}

\begin{definition}[TR model (\cite{zhao2016tensor})]
\label{def_4}
Given $N$-th-order tensor $\mathcal{Y} = [y_{i_1,\ldots,i_N}] \in \Real^{I_1 \times \ldots \times I_N}$, its TR decomposition model, referred to as $\mathcal{Y} = {\tt TR}(\mathcal{G}^{(1)}, \mathcal{G}^{(2)}, \ldots, \mathcal{G}^{(N)})$, can be represented by circular contractions over a sequence of 3-rd order core tensors $\{\mathcal{G}^{(1)}, \mathcal{G}^{(2)}, \ldots, \mathcal{G}^{(N)} \}$, where $\mathcal{G}^{(n)} = [g_{r_n,i_n,r_{n+1}}^{(n)}] \in \Real^{R_{n} \times I_n \times R_{n+1}}$ for $n = 1, \ldots, N$ is the core tensor that captures the 2D features of $\mathcal{Y}$ across its $n$-th mode. Set $\mathcal{R} = \{R_1, \ldots, R_N \}$ denote the TR ranks, where $R_1 = R_{N+1}$. The TR model of $\mathcal{Y}$ in the element representation takes the form:
\be \label{eq_TR} 
y_{i_1,\ldots,i_N} = \sum_{n = 1}^N \sum_{r_n = 1}^{R_n} \prod_{k = 1}^N g_{r_{k},i_k,r_{k+1}}^{(k)},  
\ee
where $\forall n: r_n = 1, \ldots, R_n$ and $r_{n+1} = r_1$. 
\end{definition}

\begin{theorem}[Circular dimensional permutation invariance (\cite{zhao2016tensor})] \label{theorem}
Let $\mathcal{Y} = {\tt TR}(\mathcal{G}^{(1)}, \mathcal{G}^{(2)}, \ldots, \mathcal{G}^{(N)})$ be the TR format of $\mathcal{Y} \in \Real^{I_1 \times \ldots \times I_N}$, and let $\mathcal{Y}^{\tau_k} \in \Real^{I_{k+1} \times \ldots \times I_N \times I_1 \times \ldots \times I_k}$ express the circular shift of $\mathcal{Y}$ to the left by $k$ steps (see Definition \ref{def_3}). Then,  $\mathcal{Y}^{\tau_k}  = {\tt TR}(\mathcal{G}^{(k+1)}, \ldots, \mathcal{G}^{(N)}, \mathcal{G}^{(1)}, \ldots, \mathcal{G}^{(k)})$.
 \end{theorem}

\begin{definition}[2D Convolution]
Let $\mathcal{X} =  [x_{i_1,i_2,c}] \in \Real^{I_1 \times I_2 \times C}$ be the input activation maps, $\mathcal{Y} = [y_{\tilde{i}_1,\tilde{i}_2,t}] \in \Real^{\tilde{I}_1 \times \tilde{I}_2 \times T}$ be the output feature maps, and $\mathcal{W} = [w_{t,c,d_1,d_2}] \in \Real^{T \times C \times D_1 \times D_2}$ be the convolutional kernel. Then the 2D convolution maps input tensor $\mathcal{X}$ to output tensor $\mathcal{Y}$ by the following linear transformation:
\be \label{conv_kr}  y_{\tilde{i}_1,\tilde{i}_2,t} = \sum_{c = 1}^C \sum_{d_1 = 1}^{D_1} \sum_{d_2 = 1}^{D_2} w_{t,c,d_1,d_2} x_{i_1(d_1), i_2(d_2),c}, \ee
where $i_1(d_1) = (\tilde{i}_1 - 1)\Delta + d_1 - P$,  $i_2(d_2) = (\tilde{i}_2 - 1)\Delta + d_2 - P$, $\tilde{i}_1 = 1, \ldots, \frac{I_1 + 2P - D_1}{\Delta} +1$, $\tilde{i}_2 = 1, \ldots, \frac{I_2 + 2P - D_2}{\Delta} +1$, $\Delta$ is the stride, and $P$ is the zero-padding size.
Using the simplified tensor operator notation, mapping (\ref{conv_kr}) can be equivalently rewritten as:
\be \label{conv_kr_1}  \mathcal{Y} = \mathcal{X} \times_3^2 \star_{\{1, 2 \}}^{\{3,4\}}\mathcal{W}, \ee
where $\times_3^2$ means that the third mode of $\mathcal{X}$ is contracted with the second mode of $\mathcal{W}$, and symbol $\star_{\{1, 2 \}}^{\{3,4\}}$ denotes that the first two modes of $\mathcal{X}$ are convoluted with the last two modes of $\mathcal{W}$. \textbf{} 
    
\end{definition}

\section{Proposed method}
\label{sec:proposed_method}
\begin{figure*}
    \centering
\includegraphics[scale=0.73]{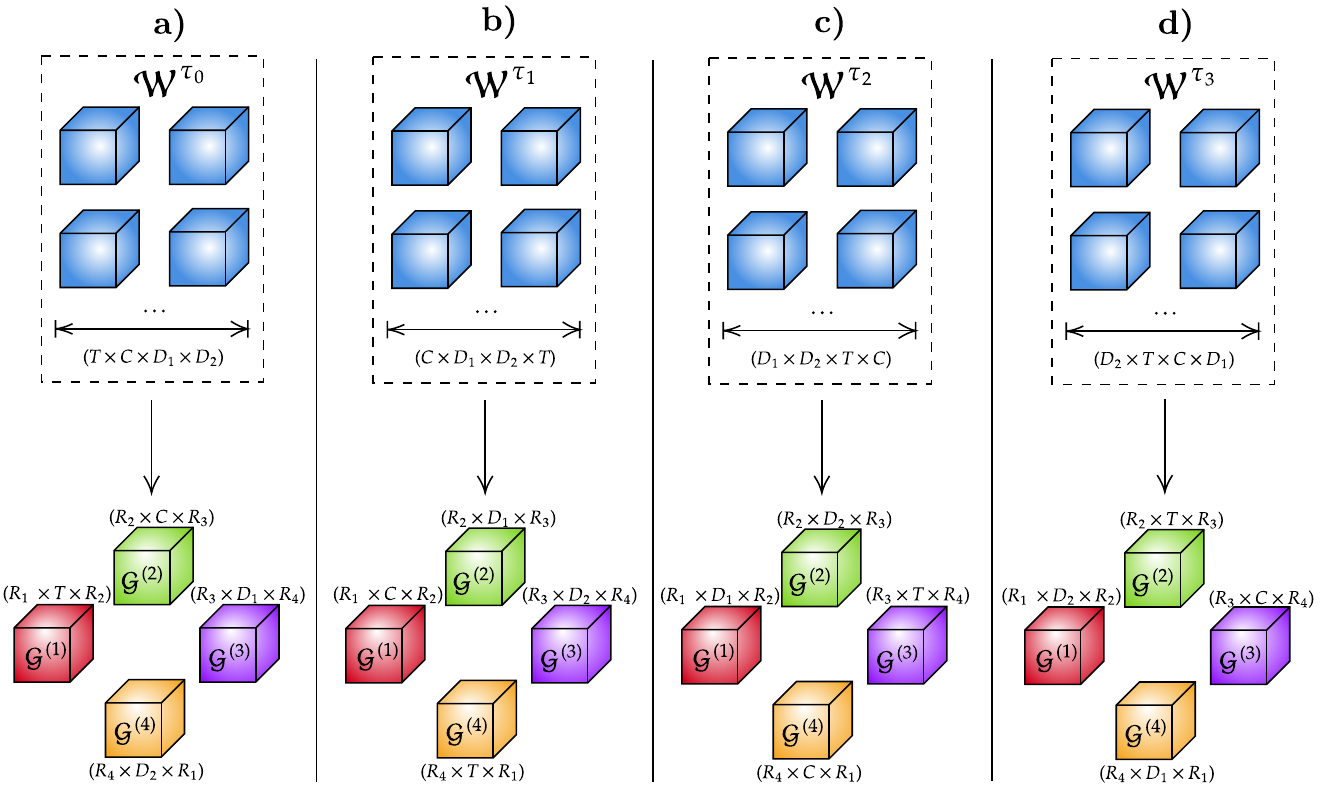}
    \caption{TR decompositions of all circular mode-permutations of the kernel weight tensor.}
    \label{fig:TR_permutations}
\end{figure*}

The weight tensor of the convolutional layer is represented as 4-th-order tensor $\mathcal{W} = [w_{t,c,d_1,d_2}] \in \Real^{T \times C \times D_1 \times D_2}$, where $C$ and $T$ represent the input and output channels, respectively, and $D_1, D_2$ represent the height and width of the filter, respectively. The original weight tensor $\mathcal{W}$ in the TR model (Definition \ref{def_4}) is approximated as follows:
\begin{equation}\label{perm1}
    \mathcal{W}  \approx \mathcal{G}^{(1)} \times_3^1 \mathcal{G}^{(2)} \times_4^1  \mathcal{G}^{(3)} \times_{5,1}^{1,3}  \mathcal{G}^{(4)},
\end{equation}
where $\mathcal{G}^{(1)} \in \Real^{R_1 \times T \times R_2}$, $\mathcal{G}^{(2)} \in \Real^{R_2 \times C \times R_3}$, $\mathcal{G}^{(3)} \in \Real^{R_3 \times D_1 \times R_4}$, and $\mathcal{G}^{(4)} \in \Real^{R_4 \times D_2 \times R_1}$. The weight tensor in (\ref{perm1}) is unpermuted and the circular shift (Definition \ref{def_3}) is $\tau_0$. However, as noted by \cite{mickelin2020algorithms}, due to the circular-dimensional permutation invariance property of TR (Theorem \ref{theorem}), we can find a TR model with much lower storage cost by decomposing the tensor on different available circular permutations. All circular permutations of the original weight tensor $\mathcal{W}$ are visualized in Figure \ref{fig:TR_permutations}.

Inserting model (\ref{perm1}) into mapping (\ref{conv_kr}), separating 2D convolution into 1D convolutions, rearranging the summands, and using  tensor operator notations, the mapping can be simplified as follows: 
\begin{equation}\label{conv_kr_tr}
    \begin{split}
        \mathcal{Y} & = \mathcal{X} \times_3^2 \star_{\{1, 2 \}}^{\{3,4\}}\mathcal{W} \\
        & = \left \{ \left [ \underbrace{\left ( \textcolor{blue}{\mathcal{X} \times_3^2 \mathcal{G}^{(2)}}  \right )}_\text{\textcolor{blue}{contraction}}  \times_4^1 \star_{\{1 \}}^{\{2 \}}  \mathcal{G}^{(3)}  \right ] \times_4^1 \star_{\{2 \}}^{\{2 \}}  \mathcal{G}^{(4)}  \right \} \times_{4,3}^{1,3} \mathcal{G}^{(1)} \\
        & = \left \{ \underbrace{ \left [ \textcolor{red}{\mathcal{Z}  \times_4^1 \star_{\{1 \}}^{\{2 \}}  \mathcal{G}^{(3)}}  \right ]}_\text{\textcolor{red}{$D_1 \times 1$ conv.}} \times_4^1 \star_{\{2 \}}^{\{2 \}}  \mathcal{G}^{(4)}  \right \} \times_{4,3}^{1,3} \mathcal{G}^{(1)} \\
        & = \underbrace{\left \{ \color{teal}{\mathcal{Z}^{(V)} \times_4^1 \star_{\{2 \}}^{\{2 \}}  \mathcal{G}^{(4)}}  \right \}}_\text{\textcolor{teal}{$1 \times D_2$ conv.}} \times_{4,3}^{1,3} \mathcal{G}^{(1)} \\
        & = \underbrace{\color{violet}{\mathcal{Z}^{(V,H)}  \times_{4,3}^{1,3}  \mathcal{G}^{(1)} }}_\text{\textcolor{violet}{contraction}} 
    \end{split}
\end{equation}

\begin{figure*}[!ht]
    \centering
    \includegraphics[scale=0.86]{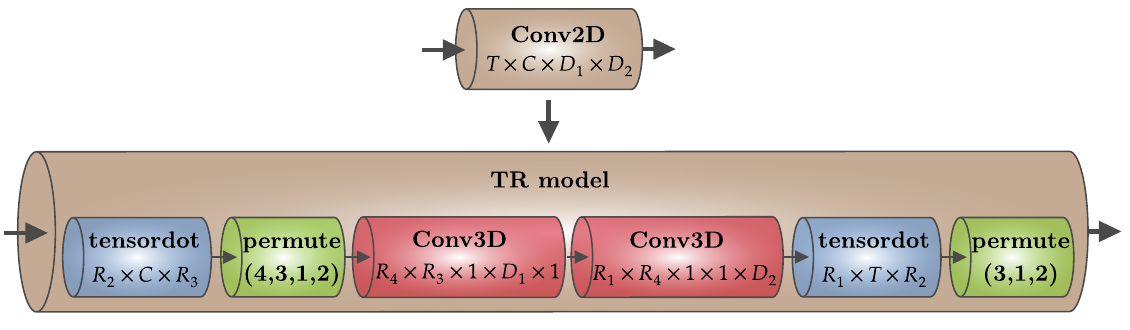}
    \caption{Visual representation of the proposed new layer constructed from the decomposed factors in  \textit{Pytorch} library for circular shift $\tau_0$.}
    \label{fig:dtr}
\end{figure*}

Mapping (\ref{conv_kr_tr}) in a convolutional layer can thus be regarded as a pipeline of fundamental tensor computations performed on much smaller data blocks. Physically, to implement all pipeline operations, each convolutional layer is replaced with a four-sublayer structure, which is demonstrated in Fig. \ref{fig:dtr}. The first sublayer performs the contraction operation involved in the computation of $\mathcal{Z}$, which has a similar functionality as a standard fully-connected layer. Thus, mapping 
\be \label{conv_kr_tr_1} \mathcal{Z} = \mathcal{X} \times_3^2 \mathcal{G}^{(2)} \in \Real^{I_1 \times I_2 \times R_2 \times R_3}
\ee 
can be performed with the {\tt tensordot} function in PyTorch. The activation tensor $\mathcal{Z}$ computed in the first sublayer is then provided (after permutation) to the second convolutional sublayer represented by $\mathcal{G}^{(3)} \in \Real^{R_3 \times D_1 \times R_4}$, which is much smaller than $\mathcal{W}$, and this sublayer computes the 1D convolutions along the 1-st mode (vertically):
 \be \label{eq_Z_V} \mathcal{Z}^{(V)}  = \mathcal{Z} \times_4^1 \star_{\{1 \}}^{\{2 \}}  \mathcal{G}^{(3)} \in \Real^{\tilde{I}_1 \times I_2 \times R_2 \times R_4}.
 \ee
 The 1D convolution on a 4D data in $\mathcal{Z}$ can be implemented using the {\tt Conv3D} class of PyTorch.  Due to its syntax rules, the output from the first sublayer needs to be mode-permuted (the green block on the left), and the weights in $\mathcal{G}^{(3)}$ should also be mode-permuted and singular-mode extended as follows: $\Real^{R_3 \times D_1 \times R_4} \ni \mathcal{G}^{(3)} \rightarrow  \tilde{\mathcal{G}}^{(3)}  \in \Real^{R_4 \times R_3 \times 1 \times D_1 \times 1}$.   Next, the third convolutional sublayer is created to compute the 1D convolutions along the horizontal direction of activation tensor $\mathcal{Z}^{(V)}$. The output activation tensor obtained from this sublayer has the form:  
   \be \label{eq_Z_VH} \mathcal{Z}^{(V,H)}  = \mathcal{Z}^{(V)} \times_4^1 \star_{\{2 \}}^{\{2 \}}  \mathcal{G}^{(4)} \in \Real^{\tilde{I}_1 \times \tilde{I}_2 \times R_2 \times R_1}.
 \ee
  This convolution operation can also be executed with the {\tt Conv3D} class that involves a repermuted and extended core tensor $\mathcal{G}^{(4)} \rightarrow \tilde{\mathcal{G}}^{(4)} \in \Real^{R_1 \times R_4 \times 1 \times 1 \times D_2}$. 
  Finally, the fourth sublayer is created, which performs two-mode contraction operations according to the model: 
 \be \label{eq_Y_4} \mathcal{Y} = \mathcal{Z}^{(V,H)} \times_{4,3}^{1,3} \mathcal{G}^{(1)} \in \Real^{\tilde{I}_1 \times \tilde{I}_2 \times T}. \ee

The final version of the TR-convolution algorithm is listed in Algorithm \ref{alg1b}.
\begin{algorithm}[!ht]
\caption{\bf TR-convolution} \label{alg1b}
\SetKwInOut{Input}{Input} \SetKwInOut{Output}{Output}
\Input{$\mathcal{X} \in \Real^{I_1 \times I_2 \times C}$ -- input activation tensor, \\ $\{{\mathcal{G}}^{(1)}, \ldots, {\mathcal{G}}^{(4)} \}$ -- TR core tensors}
\Output{$\mathcal{Y} \in \Real^{\tilde{I}_1 \times \tilde{I}_2 \times T}$ -- output activation tensor}
 \BlankLine
 Compute $\mathcal{Z}$ as in (\ref{conv_kr_tr_1}) using the $1 \times 1$ convolution,
 
 Compute $\mathcal{Z}^{(V)}$ as in (\ref{eq_Z_V}) using the 1D conv. along the vertical direction,
 
 Compute $\mathcal{Z}^{(V,H)}$ as in (\ref{eq_Z_VH}) using the 1D conv. along the horizontal direction, 
 
 Compute $\mathcal{Y}$ according to (\ref{eq_Y_4}) using the two-mode contraction operation,
 \end{algorithm}

The derivations of convolution for TR representations in other circular mode-permutations are similar, and for the sake of clarity, are explained in \ref{appendix_convolutions}.

Although TR is a circular-dimensional permutation-invariant model (\cite{zhao2016tensor}), the ordering of modes considerably affects the storage complexity, especially as the kernel weight tensors are strongly mode-unbalanced. To analyze how a storage complexity depends on a circular mode-permutation and the choice of $R_1$ (first-mode rank), we present the upper bounds of the storage complexity for each circular permutation case in \ref{appendix:storage_complex}. 

The total number of FLOPS for the convolution in (\ref{conv_kr_tr}) is calculated as follows. The first contraction in (\ref{conv_kr_tr_1}) and the last contraction in (\ref{eq_Y_4}) take $\mathcal{O}(R_2CR_3I_1I_2)$ and $\mathcal{O}(R_1TR_2\tilde{I_1}\tilde{I_2})$ FLOPS. The horizontal convolution in (\ref{eq_Z_V}) and vertical convolution in (\ref{eq_Z_VH}) are bounded by $\mathcal{O}(R_3D_1R_4R_2I_1I_2)$ and $\mathcal{O}(R_4D_2R_1R_2\tilde{I_1}I_2$, respectively. The total number of FLOPS of the entire convolution in (\ref{conv_kr_tr}) can be estimated as $\mathcal{O}(R_2CR_3I_1I_2 + R_3D_1R_4R_2I_1I_2 + R_4D_2R_1R_2\tilde{I_1}I_2 + R_1TR_2\tilde{I_1}\tilde{I_2})$. For convolutions in other circular shifts the calculations are similar, and the complexities for each case are shown in Table \ref{tab:storage_complexities}.

\begin{table*}[!ht]
\centering
\caption{Storage and FLOPS complexities for all circular mode-permutations of the kernel weight tensor.}
\label{tab:storage_complexities}
\begin{tabular}{@{}ccc@{}}
\toprule
Perm.& Storage complexity & FLOPS complexity \\ \midrule
 $\tau_0$ & $\mathcal{O}(R_2CR_3 + R_3D_1R_4 + R_4D_2R_1 + R_1TR_2)$ & $\mathcal{O}(R_2CR_3I_1I_2 + R_3D_1R_4R_2I_1I_2 + R_4D_2R_1R_2\tilde{I_1}I_2 + R_1TR_2\tilde{I_1}\tilde{I_2})$ \\
 $\tau_1$ & $\mathcal{O}(R_1CR_2 + R_2D_1R_3 + R_3D_2R_4 + R_4TR_1)$ & $\mathcal{O}(R_1CR_2I_1I_2 + R_2D_1R_3R_1I_1I_2 + R_3D_2R_4R_1\tilde{I_1}I_2 + R_4TR_1\tilde{I_1}\tilde{I_2})$  \\
 $\tau_2$ & $\mathcal{O}(R_4CR_1 + R_1D_1R_2 + R_2D_2R_3 + R_4TR_4)$ & $\mathcal{O}(R_4CR_1I_1I_2 + R_1D_1R_2R_4I_1I_2 + R_2D_2R_3R_4\tilde{I_1}I_2 + R_3TR_4\tilde{I_1}\tilde{I_2})$  \\
 $\tau_3$ & $\mathcal{O}(R_3CR_4 + R_4D_1R_1 + R_1D_2R_2 + R_2TR_3)$ & $\mathcal{O}(R_3CR_4I_1I_2 + R_4D_1R_1R_3I_1I_2 + R_1D_2R_2R_3\tilde{I_1}I_2 + R_2TR_3\tilde{I_1}\tilde{I_2})$ \\ \bottomrule
\end{tabular}
\end{table*}

\begin{remark}
    For simplicity of analysis, assume that $D=D_1=D_2$, and let all the ranks in the TR model of convolutional weight tensor $\mathcal{W}$ in (\ref{perm1}) to be equal: $R = R_1=R_2=R_3=R_4$. In such a case, all circular mode-permuted versions of the TR model simplify to one case, where the FLOPS complexity is $\mathcal{O}(R^2CI_1I_2 + R^3DI_1I_2 + R^3D\tilde{I}_1I_2 + R^2T\tilde{I}_1\tilde{I}_2)$. Comparing it with the FLOPS complexity of the tensorized TR approach (see \ref{flops_tensorized}) and removing the same terms, we can define the following FLOPS complexity ratio:
    \begin{equation}\label{ratio}
        \rho = \frac{RDI_1I_2 + RD\tilde{I_1}I_2}{R(J_1J_2+ C + T + O_1O_2) + D^2I_1I_2},
    \end{equation}
    where $J_1, J_2$ and $O_1,O_2$ are the first two dimensions of tensorized input ($C=J_1J_2J_3$) and output ($T=O_1O_2O_3$) feature maps in the tensorized TR (\cite{wang2018wide}), respectively (see \ref{appendix_tr_models}).
    
    At first glance, it is not easy to notice when $\rho$ in (\ref{ratio}) is greater than one, because $I_1,I_2, \tilde{I_1},C,T, J_1, J_2, O_1, O_2$ change in subsequent layers in CNNs. Hence, let us consider all possible examples of layers in ResNet-32. The real parameters to calculate $\rho$ for each type of layer ($L_1$, $L_2$, \dots) are shown in Table \ref{tab:layers_raio}.

    \begin{table}[!ht]
    \centering
    \caption{Parameters of layers to compute $\rho$.}
    \label{tab:layers_raio}
    \begin{tabular}{cccccc}
    \hline
    Parameter & $L_1$ & $L_2$ & $L_3$ & $L_4$ & $L_5$ \\ \hline
    $I_1$ & 32 & 32 & 16 & 16 & 8 \\
    $I_2$ & 32 & 32 & 16 & 16 & 8 \\
    $\tilde{I_1}$ & 32 & 16 & 16 & 8 & 8 \\
    $C$ & 16 & 16 & 32 & 32 & 64 \\
    $T$ & 16 & 32 & 32 & 64 & 64 \\
    $D$ & 3 & 3 & 3 & 3 & 3 \\
    $J_1$ & 4 & 4 & 4 & 4 & 4 \\
    $J_2$ & 2 & 4 & 4 & 4 & 4 \\
    $O_1$ & 4 & 4 & 4 & 4 & 4 \\
    $O_2$ & 2 & 4 & 4 & 4 & 4 \\ \hline
    \end{tabular}
    \end{table}

    Based on the real parameters in Table \ref{tab:layers_raio}, we calculated $\rho$ with rank $R$ in the range of $[1,30]$, which is shown in Figure \ref{fig:rank_ratio}.
    \begin{figure}[!ht]
        \centering
        \includegraphics[scale=0.52]{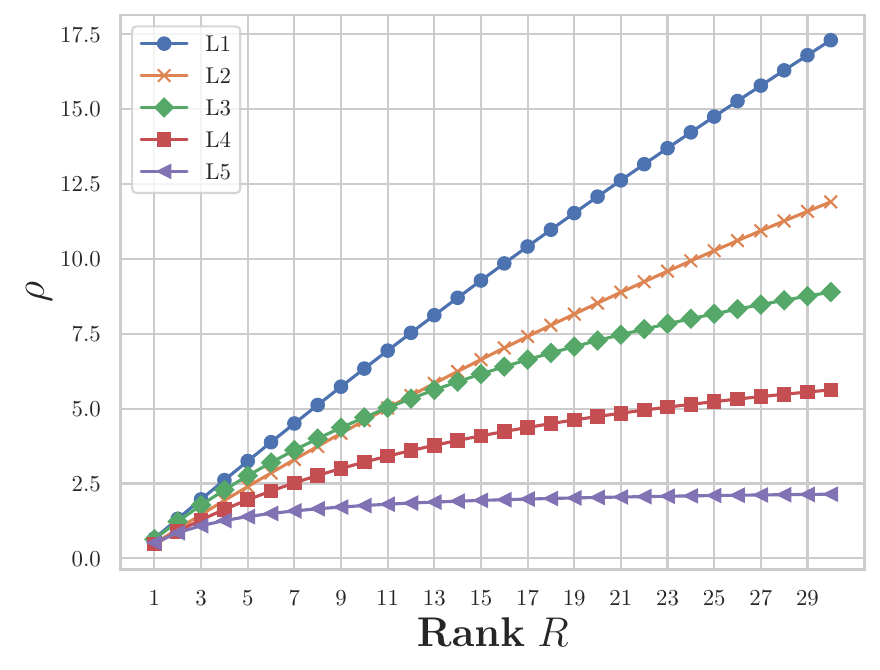}
        \caption{Ratio $\rho$ versus rank $R$ in the range of $[1,30]$ for all types of convolutional layers in the ResNet-32 network..}
        \label{fig:rank_ratio}
    \end{figure}
    Layer $L_1$ exists in the first 5 ResNet blocks in ResNet-32. Layer $L_2$ is a downsampling layer that occurs once after five ResNet blocks of layer $L_1$, and similarly to $L_1$, $\rho$ grows with a higher rank. Layer $L_3$ refers to four ResNet blocks which follow  layer $L_2$. $L_4$ similarly to $L_2$ is a downsampling layer that occurs only once. The network ends with four ResNet blocks with layer $L_5$. As can be seen, we have $\rho > 1$ for each case, which shows the superiority of the proposed method compared to the tensorized version of TR. The effect of larger FLOPS compression can be seen for the first layers ($L_1$, $L_2$) of ResNet, and is lower for the deeper layers ($L_3$, $L_4$, $L_5$).        
\end{remark}

\section{Experiments}
\label{sec:experiments}
The proposed method, known as reduced storage direct tensor ring decomposition (RSDTR), has been extensively tested in conjunction with various CNN architectures that differ in size and number of layers. RSDTR\footnote{The \textit{Pytorch} source code of the RSDTR method with the examples of usage are available at the following link: \url{https://github.com/mateuszgabor/rsdtr_compression}}  was implemented as the reduced storage TR-SVD where the optimal circular mode-permutation and divisor $R_1$ are found by the exhaustive search procedure (Algorithm 2 in \cite{mickelin2020algorithms}). The tests were carried out on two image classification datasets: CIFAR-10 and ImageNet.

\subsection{Experimental setup}
\label{sec:exp_setup}
Using CIFAR-10, we trained three ResNet networks (\cite{he2016deep}): ResNet-20, ResNet-32, ResNet-56, and VGG-19 network (\cite{simonyan2014very}). The baseline networks were trained according to the guidelines in the original paper of ResNets (\cite{he2016deep}). The networks were trained for 200 epochs using the SGD algorithm with a momentum of 0.9, and a mini-batch size equal to 128. We set the weight decay to $10^{-4}$ and the initial learning rate to 0.1, which was decayed by a factor of 0.1 after 100 and 150 epochs. 

After compressing baseline models by RSDTR, compressed networks were fine-tuned for 160 epochs with SGD without momentum with the learning rate of 0.01 and the weight decay of $10^{-3}$. In the fine-tuning proces, we used the \textit{ one-cycle learning rate} policy. FLOPS for tensorized TR networks were calculated according to their implementation in the \textit{Tednet} library (\cite{pan2022tednet}) and the ranks reported in the original papers.

\begin{figure}[!ht]
    \centering
    \includegraphics[scale=0.27]{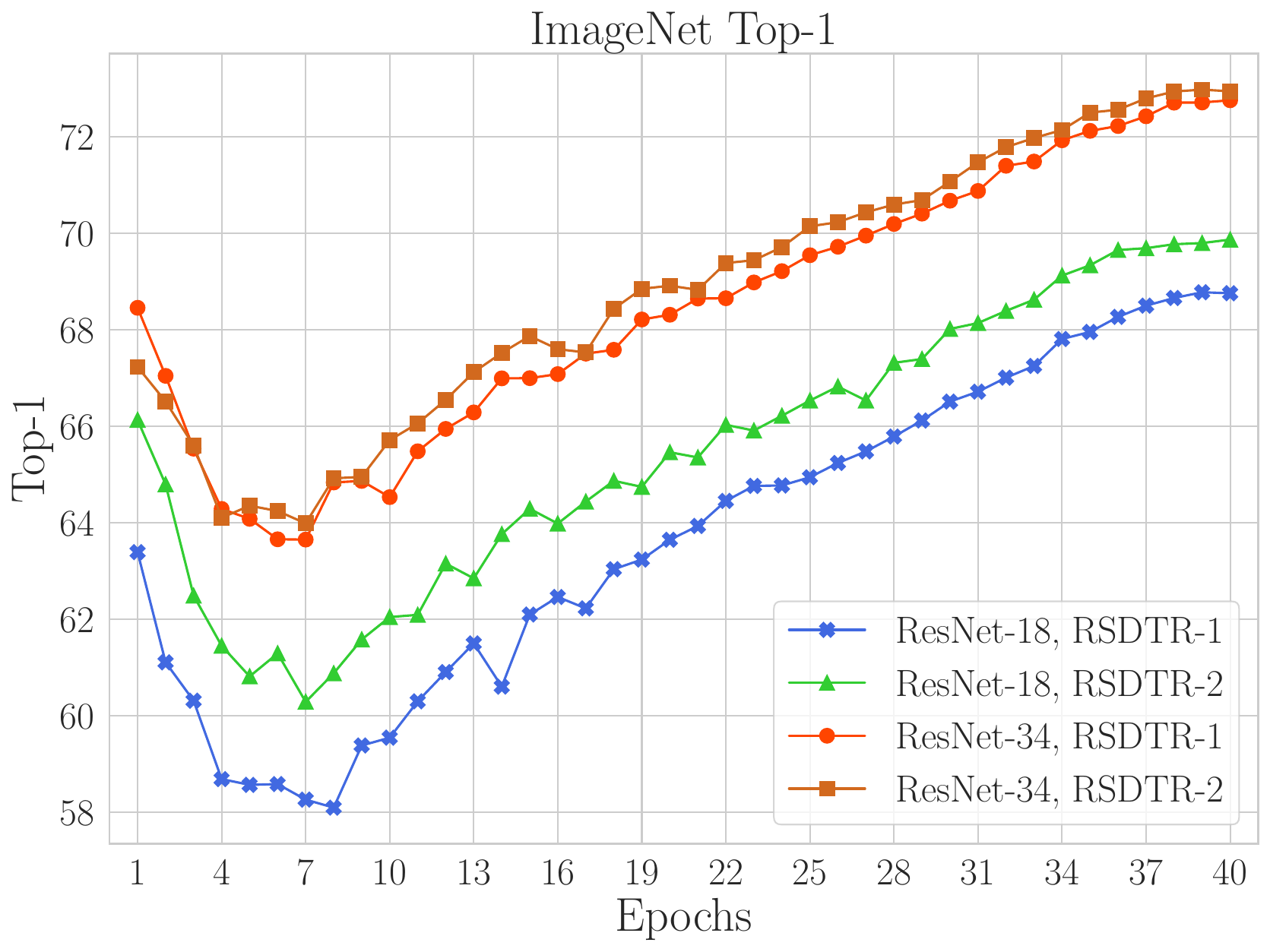}
    \caption{Fine-tuning curves for compressed ResNet-18 and ResNet-34 by RSDTR method analyzed on the ImageNet dataset.}
    \label{fig:imagenet_fine_tuning}
\end{figure}

To evaluate our method on the ImageNet dataset, we used the ResNet-18 and ResNet-34 networks. The pretrained models were downloaded from the \textit{PyTorch} package \textit{Torchvision}. The compressed networks were fine-tuned for 40 epochs using the SGD algorithm with a momentum of 0.9. The batch size was equal to 128, the learning rate was 0.01, and the weight decay was set to zero. The fine-tuning curves obtained for RSDTR on the ImageNet dataset are shown in Figure \ref{fig:imagenet_fine_tuning}.

For both datasets, the same augmentation technique was used as in the original paper on ResNet (\cite{he2016deep}). Table \ref{tab:baselines} shows the total number of parameters and FLOPS of all used CNNs.
\begin{table}[!ht]
\caption{Total number of parameters and FLOPS for baseline networks}
\centering
\begin{tabular}{@{}lll@{}}
\toprule
Network           & Params & FLOPS   \\ \midrule
ResNet-20         & 270K    & 40.55M  \\
ResNet-32         & 464K    & 68.86M  \\
ResNet-56         & 853K    & 125M \\
VGG-19             & 20.2M & 398M \\
ResNet-18         & 11.7M  & 1.81G   \\
ResNet-34         & 21.8M  & 3.66G   \\
\bottomrule
\end{tabular}
\label{tab:baselines}
\end{table}

\subsubsection{Compression scheme}
The RSDTR-based compression experiments were carried out according to the following scheme: 
\begin{enumerate}
    \item In the first stage, each convolutional kernel (except the first one) is decomposed using the RSDTR method.
    \item In the second step, the decomposed factors are used as new initial weights, which replace the original convolutional layer with a pipeline of two contractions and two 1D convolutions.
    \item Finally, each compressed network is fine-tuned to achieve an accuracy similar to the baseline model.
\end{enumerate}
RSDTR was tested with different levels of compression, controlled by the prescribed relative error $\epsilon_p$ in the TR algorithm. The prescribed relative errors for each tested case are given in parentheses in the tables/figures below. 

The experiments with CIFAR-10 were run on the single NVIDIA RTX 3080 Ti GPU and on the Google Colab platform  In the case of the ImageNet dataset, the experiments were performed on the Amazon Elastic Compute Cloud (EC2) p3.8.large instance from Amazon Web Services.

\subsubsection{Evaluation metrics}
To evaluate the compression of neural networks, two compression metrics are used, measuring the parameter and FLOPS compression. The parameter compression ratio (PCR) is defined as: 
\begin{equation}
    {\Downarrow}{\textrm{Params}} = \frac{\textrm{Params(baseline network)}}{\textrm{Params(compressed network)}}
\end{equation}
 which compares the total number of parameters of the baseline network to the compressed network. The FLOPS compression ratio (FCR) is defined in a similar way: 
 \begin{equation}
     {\Downarrow}\textrm{FLOPS} = \frac{\textrm{FLOPS(baseline network)}}{\textrm{FLOPS(compressed network)}}.
 \end{equation}

The drop in classification accuracy compared to the accuracy of the baseline network reported in a given paper is marked as ${\Delta}$top-1 and is given in percentages. 

\subsection{Results}
\label{sec:results}
In this section, the results of compression of each network will be shown compared with state-of-the-art competitive methods.

\begin{figure*}[!ht]
    \centering
\includegraphics[scale=0.265]{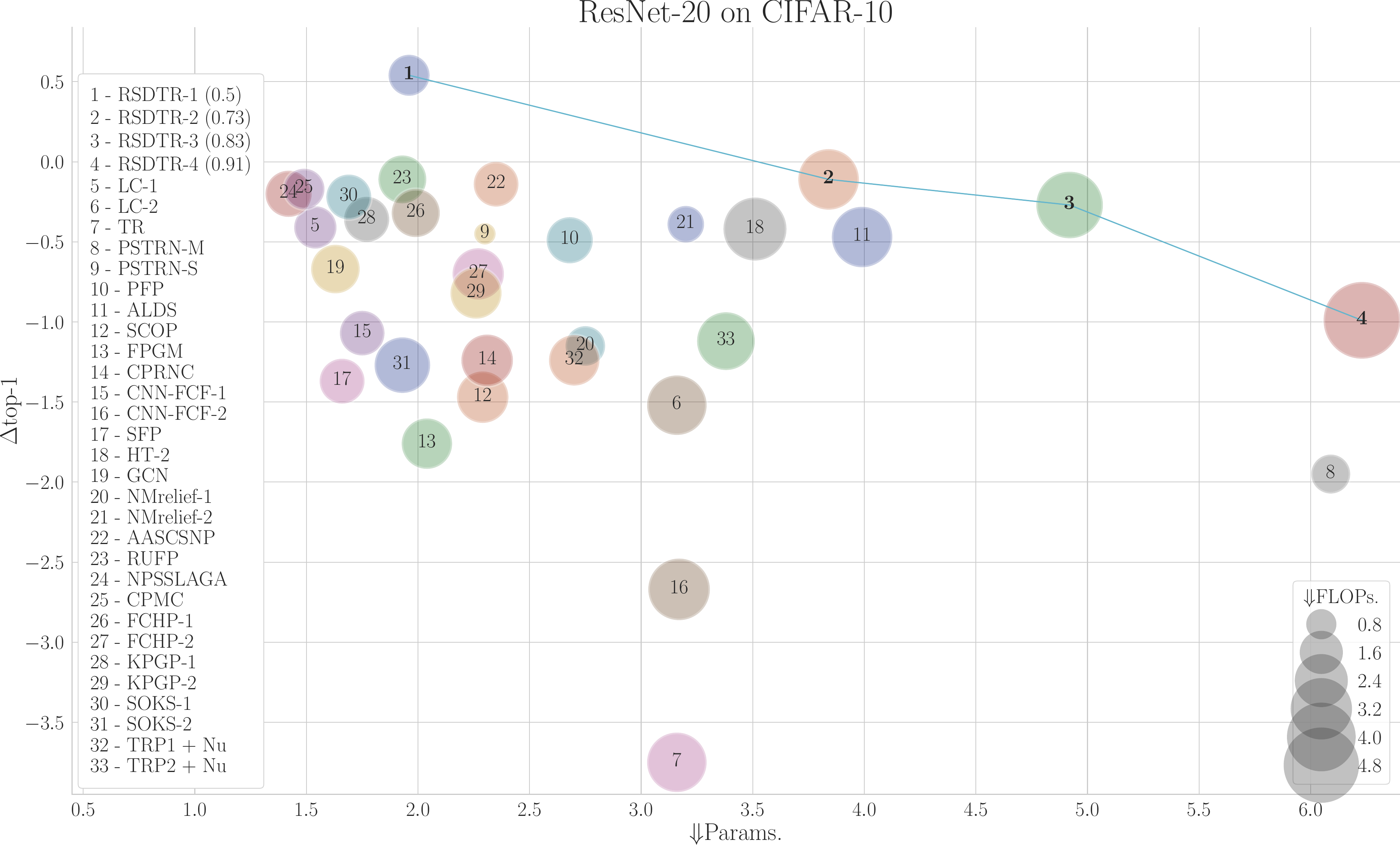}
    \caption{Results of ResNet-20 compression on CIFAR-10.}
    \label{fig:cifar10_resnet20}
\end{figure*}
\subsubsection{ResNet-20}
Figure \ref{fig:cifar10_resnet20} presents the results obtained with RSDTR and other compression methods that were applied to the ResNet-20 network. Comparing RSDTR with pruning approaches: PFP (\cite{liebenwein2020provable}), ALDS (\cite{liebenwein2021compressing}), SCOP (\cite{tang2020scop}), CPRNC (\cite{wu2024cprnc}), CNN-FCF (\cite{li2019compressing}), SFP (\cite{he2018soft}), GCN (\cite{di2022channel}), NNrelief (\cite{dekhovich2024neural}), AASCSNP (\cite{liu2024attention}), RUFP  
(\cite{zhang2022rufp}), NPSSLAGA (\cite{wang2020network}), CPMC (\cite{yan2021channel}), FCHP (\cite{zhang2022fchp}), KPGP (\cite{zhang2022group}), SOKS (\cite{liu2022soks}), TRP (\cite{xu2019trained}), it can be seen that RSDTR outperforms all compared pruning approaches in terms of PCR, FCR, and the drop in classification accuracy. Furthermore, with reference to the low-rank methods: TR (\cite{wang2018wide}), LC (\cite{idelbayev2020low}), PSTRN (\cite{li2021heuristic}), HT-2 (\cite{gabor2023compressing}), RSDTR achieved a higher accuracy for a similar value of PCR (or even higher) and a larger FCR. Interestingly, FCR of TR and PSTRN-S is 0.84 and 0.42, which means that the FLOPS number for such compressed neural networks is 1.19 and 2.38 higher than for the uncompressed model. This problem does not exist for RSDTR for which the parameter compression goes along with FLOPS compression. 

\begin{figure*}[!ht]
    \centering
\includegraphics[scale=0.265]{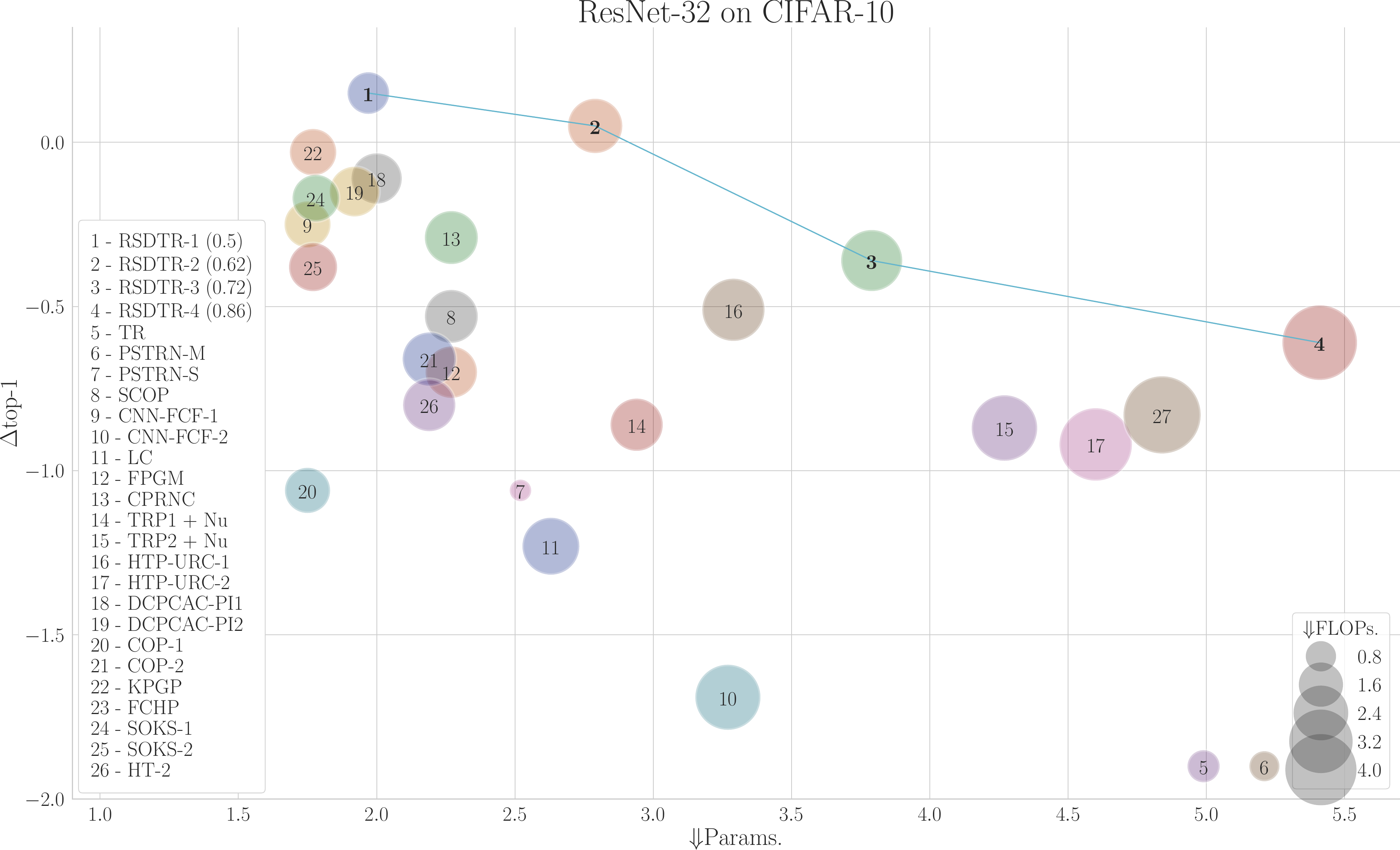}
    \caption{Results of ResNet-32 compression on CIFAR-10.}
    \label{fig:cifar10_resnet32}
\end{figure*}
\subsubsection{ResNet-32}
Similarly to ResNet-20, we carried out the experiments using a 12-layer deeper network, i.e., ResNet-32. The results given in Figure \ref{fig:cifar10_resnet32} clearly demonstrate that RSDTR achieved a higher accuracy than the tensorized versions: TR (\cite{wang2018wide}), PSTRN (\cite{li2021heuristic}), and other low-rank methods: LC (\cite{idelbayev2020low}), HT-2 (\cite{gabor2023compressing}) at a similar value of PCR. The same as for ResNet-20, FCR is lower than one for PSTRN-S (0.79), PSTRN-M (0.45) and TR (0.89), which means that the number of FLOPS increases for these approaches with respect to the uncompressed network. The pruning methods, such as: SCOP (\cite{tang2020scop}), CNN-FCF (\cite{li2019compressing}), FPGM (\cite{he2019filter}), TRP (\cite{xu2019trained}), HTP-URC (\cite{qian2023hierarchical}), DCPCAC (\cite{chen2020dynamical}), COP (\cite{wang2021cop}), FCHP (\cite{zhang2022fchp}), KPGP (\cite{zhang2022group}), SOKS (\cite{liu2022soks}) gave worse results than RSDTR for each tested case. 

\begin{figure*}[!ht]
    \centering
\includegraphics[scale=0.265]{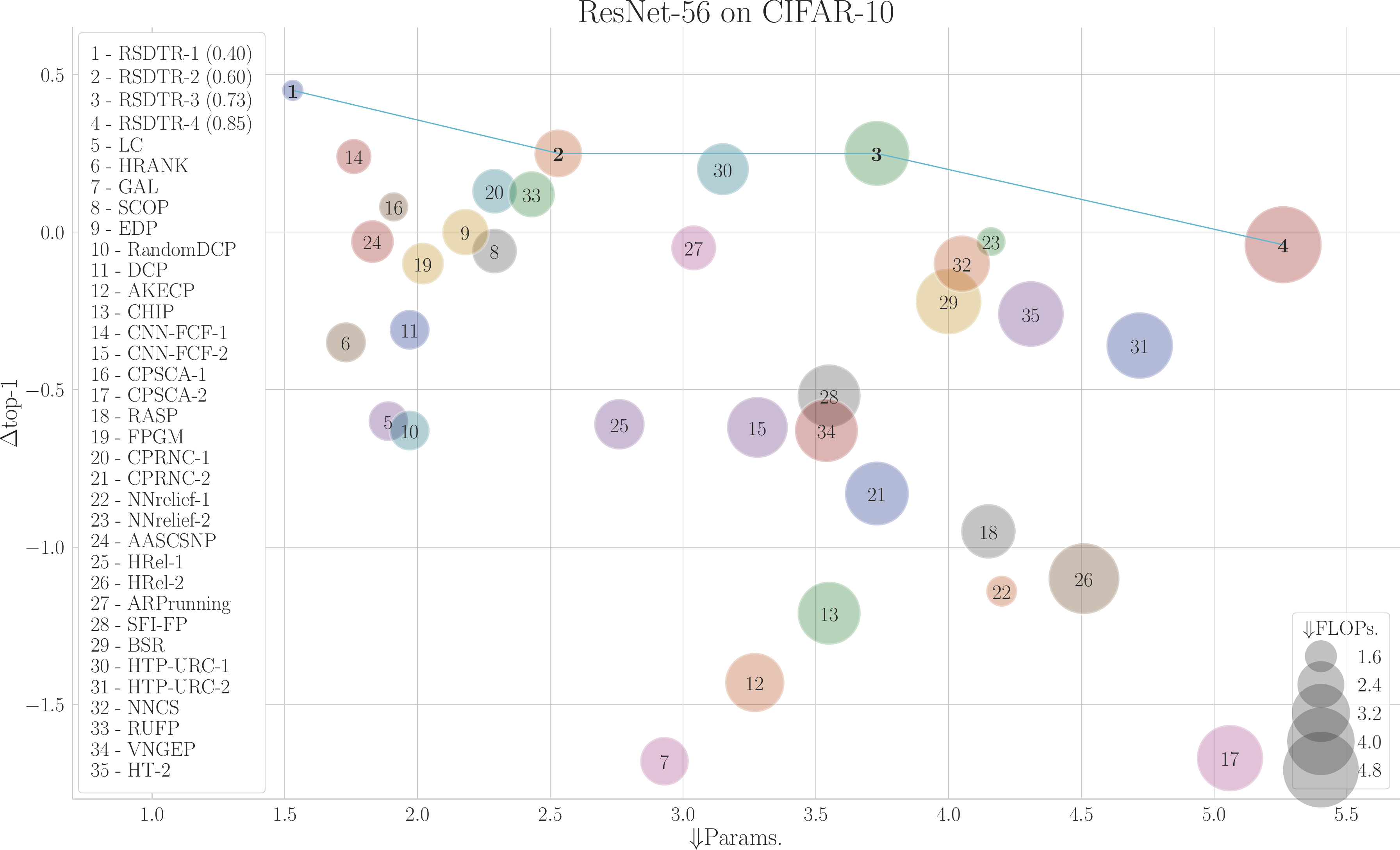}
    \caption{Results of ResNet-56 compression on CIFAR-10.}
    \label{fig:resnet56}
\end{figure*}
\subsubsection{ResNet-56}
The next network that we analyze on CIFAR-10 is ResNet-56. This neural network is one of the most extensively studied in the literature and is mainly dominated by pruning methods. Figure \ref{fig:resnet56} shows the results provided by our approach with reference to the following pruning methods: HRANK (\cite{lin2020hrank}), GAL (\cite{lin2019towards}), SCOP (\cite{tang2020scop}), EDP (\cite{ruan2020edp}), DCP (\cite{zhuang2018discrimination}), Random-DCP (\cite{zhuang2018discrimination}), AKECP (\cite{zhang2021akecp}), CHIP (\cite{sui2021chip}), CNN-FCF (\cite{li2019compressing}), CPSCA (\cite{liu2021channel}), RASP  
(\cite{zhen2023rasp}), FPGM (\cite{he2019filter}), CPRNC (\cite{wu2024cprnc}), NNrelief (\cite{dekhovich2024neural}), AASCSNP (\cite{liu2024attention}), HRel (\cite{sarvani2022hrel}), ARPruning (\cite{yuan2024arpruning}),  SFI-FP (\cite{yang2024soft}), HTP-URC (\cite{qian2023hierarchical}), NNCS (\cite{gao2022efficient}), RUFP (\cite{zhang2022rufp}), VNGEP (\cite{shi2023vngep}), and the low-rank methods, such as: LC (\cite{idelbayev2020low}), HT-2 (\cite{gabor2023compressing}), BSR (\cite{eo2023effective}). The results show that the drop in accuracy of RSDTR-4 at PCR of about 5.26 and FCR of 4.96 is only $-0.04$, which is much better compared to competitive methods with similar PCR and FCR. The other cases of RSDTR show a similar scenario.

\subsubsection{VGG-19}
Table \ref{tab:cifar10_vgg16} shows the compression results obtained with RSDTR, compared to the competitive state-of-the-art methods for the VGG-19 network. We selected the following pruning methods: Sliming (\cite{liu2017learning}), CCPrune (\cite{chen2021ccprune}), PFEC (\cite{lin2019toward}), ThiNet (\cite{molchanov2019importance}), NRE (\cite{jung2019learning}), DR (\cite{zhuang2018towards}), Random-DCP (\cite{liu2021discrimination}), DCP (\cite{liu2021discrimination}), PFGD (\cite{xu2022pruning}), AAP (\cite{zhao2023automatic}), C-OBD (\cite{wang2019eigendamage}), C-OBS (\cite{wang2019eigendamage}), Kron-OBD (\cite{wang2019eigendamage}), Kron-OBS (\cite{wang2019eigendamage}), Eigendamage (\cite{wang2019eigendamage}), GT (\cite{yu2021gate}), AGSP (\cite{wei2022automatic}), and one low-rank method LC (\cite{idelbayev2020low}). As the results show, VGG-19 is heavily overparameterized and many methods achieve a higher accuracy compared to the baseline. However, RSDTR achieves the largest parameter and FLOPS compression ratio among all the compared methods. Furthermore, the increase in accuracy compared to the uncompressed model is the highest for PFS-1. RSDTR-1 achieves nearly the same increase in accuracy as for the PFS method, but the FLOPS compression for PFS is twice lower compared to that of RSDTR.
\begin{table}[!ht]
\centering
\caption{Results of VGG-19 compression on CIFAR-10.}
\begin{tabular}{@{}lcccc@{}}
\toprule
Method & $\Delta$top-1 & $\Downarrow$Params & $\Downarrow$FLOPS \\ \midrule
Sliming & $+0.14$ & 8.71 & 2.03 \\
CCPrune & $+$0.07 & 7.59 & 1.96 \\
PFEC & $+0.15$ & 2.78 & 1.52 \\
NRE  & $-0.06$ & 13.69 & 3.09 \\
DR & $-0.27$ & 8.55 & 3.19 \\
Random-DCP & $+0.03$ & 1.93 & 2.00 \\
DCP & $+0.31$ & 1.93 & 2.00 \\
PFGD & $+0.01$ & 10.14 & 3.67 \\
AAP-P & $+0.26$ & 7.14 & 2.28 \\
AAP-F & $+0.03$ & 8.08 & 2.58 \\
C-OBD  & $-0.13$ & 5.56 & 1.62 \\
C-OBS & $-0.09$ & 4.99 & 1.53 \\
Kron-OBD & $-0.17$ & 5.10 & 1.62 \\
Kron-OBS  & $-0.08$ & 4.93 & 1.59 \\
EigenDamage & $-0.19$ & 4.58 & 1.59 \\
GT-1  & $-1.07$ & 6.77 & 2.14 \\
GT-2 & $-1.09$ & 10.77 & 2.74 \\
PFS & $+0.31$ & \NA & 2.08 \\
AGSP-1 & $+0.26$ & 1.49 & 1.16 \\
AGSP-2 & $-0.87$ & 2.91 & 2.92 \\
LC & $+0.08$ & 8.21 & 4.51 \\
RSDTR (0.40) & $+0.30$ & 7.78 & 2.27 \\
RSDTR (0.57) & $+0.08$ & 15.48 & 4.97 \\
\bottomrule
\end{tabular}
\label{tab:cifar10_vgg16}
\end{table}

\subsubsection{ResNet-18}
The proposed method was also applied to compress larger CNNs. One of them is ResNet-18, for which the following pruning methods were selected for comparison: SFP (\cite{he2018soft}), MIL (\cite{dong2017more}), FPGM (\cite{he2019filter}), TRP (\cite{xu2020trp}), PFP (\cite{liebenwein2020provable}), ASFP (\cite{he2019asymptotic}), RUFP (\cite{zhang2022rufp}), FPFS (\cite{zhang2022fpfs}), FPSNC (\cite{he2022filter}), DMPP (\cite{li2022dmpp}), FTWT (\cite{elkerdawy2022fire}), TRP (\cite{xu2019trained}), ABCPrunner (\cite{lin2020channel}), EPFS (\cite{xu2021efficient}), FPSNCWGA (\cite{he2022filter}), DCPCAC (\cite{chen2020dynamical}), MDCP (\cite{zhang2023multi}), EACP (\cite{liu2023eacp}), SRFP (\cite{cai2021softer}), ASRFP (\cite{cai2021softer}), WHC (\cite{chen2023whc}), DAIS (\cite{guan2022dais}), GFBS (\cite{wei2022automatic}), CoBaL (\cite{cho2021dynamic}). The results in Table \ref{tab:resnet18} demonstrate that RSDTR outperforms all the compared approaches in almost all the metrics. EPFS-C and FTWT have slightly larger FCR, but at the cost of a larger drop in accuracy and lower PCR. 

\begin{table}[!ht]
\centering
\caption{Results of ResNet-18 compression on ImageNet.}
\begin{tabular}{lcccc}
\toprule
Method &  $\Delta$top-1 &   $\Downarrow$Params & $\Downarrow$FLOPS \\ \midrule
SFP & $-3.18$ & \NA & 1.72 \\
MIL & $-3.65$ & \NA & 1.53 \\
FPGM & $-$1.87 &  \NA & 1.72  \\
TRP & $-$3.64 &  \NA & 1.81  \\
PFP & $-$2.36 &  1.78 & 1.41 \\
ASFP & $-2.21$ &  \NA & 1.72 \\
RUFP & $-2.68$ & 1.89 & 1.85 \\
FPFS & $-2.95$ &\NA & 1.90 \\
FPSNC & $-1.97$ & \NA & 1.72 \\
FTWT & $-2.27$ & \NA & 2.06 \\
DMPP & $-1.59$ & \NA & 1.82 \\
TRP & $-3.64$ & \NA & 1.81 \\
ABCPrunner & $-2.38$ & 1.77 & 1.81 \\
EPFS-B-1 & $-1.54$ & 1.31 & 1.41 \\
EPFS-B-2 & $-2.22$ & 1.50 & 1.85 \\
EPFS-F & $-1.94$ & 1.53 & 1.72 \\
EPFS-C & $-1.63$ & 2.13 & 2.23 \\
FPSNCWGA & $-1.97$ & \NA & 1.72 \\
DCPCAC-PI1 & $-0.43$ & 1.39 & 1.38 \\
DCPCAC-PI2 & $-0.21$ & 1.56 & 1.38 \\
MDCP & $-1.63$ & \NA & 1.94 \\
RUFP & $-2.68$ & 1.89 & 1.85 \\
EACP & $-1.91$ & 2.25 & 1.99 \\
SRFP & $-2.17$ & \NA & 1.72 \\
ASRFP & $-2.98$ & \NA & 1.72 \\
WHC & $-1.28$ & \NA & 1.72 \\
DAIS & $-2.20$ & \NA & 1.81 \\
GFBS & $-1.83$ & \NA & 1.75 \\
CoBaL & $-2.11$ & 2.03 & 1.41 \\
RSDTR-1 (0.36) & $+$0.12 & 1.85 & 1.31 & \\
RSDTR-2 (0.55) & $-$0.99& 2.99 & 2.04 \\
\bottomrule
\end{tabular}
\label{tab:resnet18}
\end{table}

\subsubsection{ResNet-34}
ResNet-34 is the last and largest CNN evaluated in this study. ResNet-34 was compared with the methods similar to previously used, that is, SFP (\cite{he2018soft}), MIL (\cite{dong2017more}), CCEP (\cite{shang2022neural}), ASFP (\cite{he2019asymptotic}), PFEC (\cite{lin2019toward}), FPGM (\cite{he2019filter}), RUFP (\cite{zhang2022rufp}), DMC (\cite{gao2020discrete}), SCOP (\cite{tang2020scop}), NISP (\cite{yu2018nisp}), MDCP (\cite{zhang2023multi}), SRFP (\cite{cai2021softer}), ASRFP (\cite{cai2021softer}), TIP (\cite{yu2020tutor}), ELC (\cite{wu2023efficient}), and Taylor (\cite{molchanov2019importance}). The results, which are listed in Table \ref{tab:resnet34}, show that RSDTR outperforms all the compared methods, reaching the largest ratio of PCR to $\Delta$top-1.

\begin{table}[!ht]
\centering
\caption{Results of ResNet-34 compression on ImageNet.}
\begin{tabular}{lcccc}
\toprule
Method &  $\Delta$top-1 &  $\Downarrow$Params & $\Downarrow$FLOPS \\ \midrule
SFP & $-2.09$ & \NA & 1.70 \\
MIL & $-0.43$ & \NA & 1.33 \\
CCEP & $-0.63$ & \NA & 1.73 \\
ASFP & $-1.39$ & \NA & 1.70 \\
PFEC & $-1.06$ & \NA & 1.32 \\
FPGM & $-1.29$ & \NA & 1.70 \\
SCOP & $-0.38$ & 1.66 & 1.64 \\
NISP & $-0.28$ & 1.37 & 1.38 \\
ABCPrunner & $-2.30$ & 2.16 & 1.69 \\
MDCP & $-1.02$ & \NA & 1.73 \\
RUFP & $-1.69$ & 1.77 & 1.74 \\
SRFP & $-2.57$ & \NA & 1.70 \\
ASRFP & $-2.53$ & \NA & 1.70 \\
TIP & $-0.67$ & \NA & 1.73 \\
ELC & $-1.07$ & \NA & 1.72 \\
Taylor & $-0.48$ & \NA & 1.30 \\
RSDTR-1 (0.33) & $-$0.31 & 1.83 & 1.18 \\
RSDTR-2 (0.45) & $-$0.54 & 2.65 & 1.73 \\

\bottomrule
\end{tabular}
\label{tab:resnet34}
\end{table}

\begin{figure*}[!ht]
    \begin{subfigure}[b]{\linewidth}
        \centering
        \includegraphics[scale=0.36]{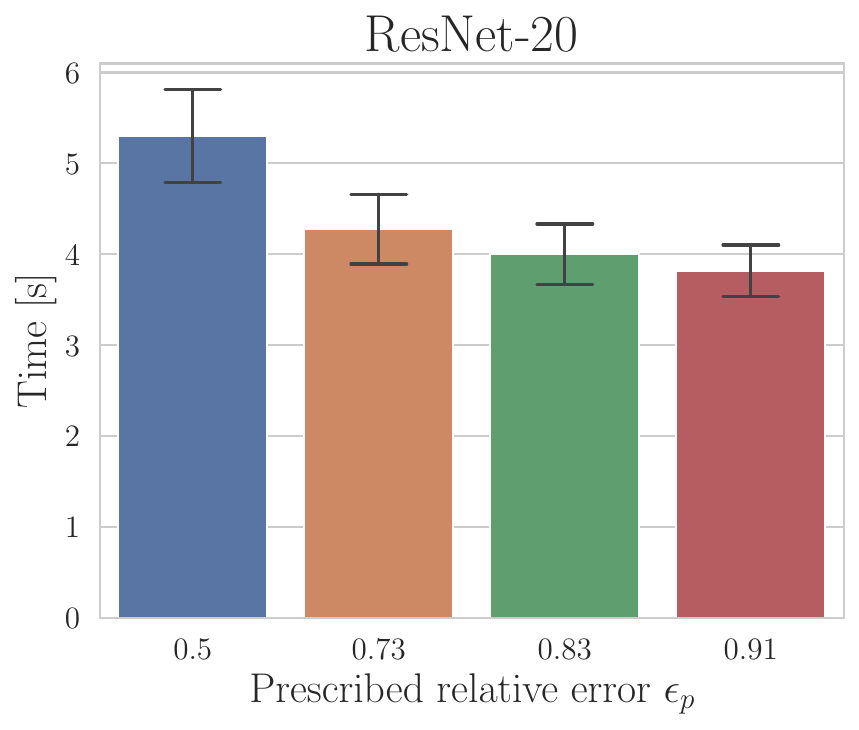}
        \hfill
        \includegraphics[scale=0.36]{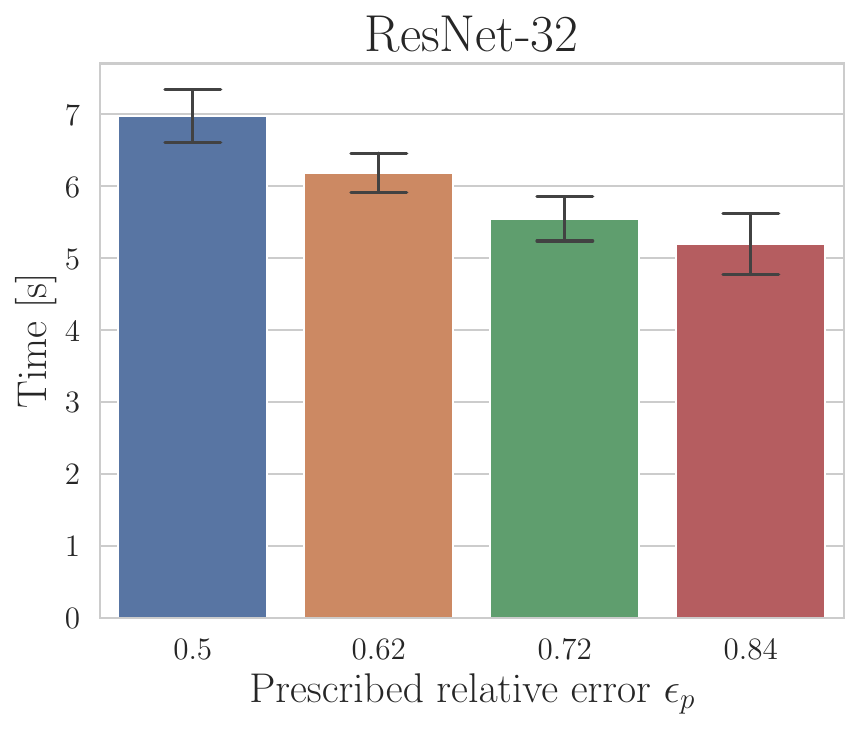}
        \hfill
        \includegraphics[scale=0.36]{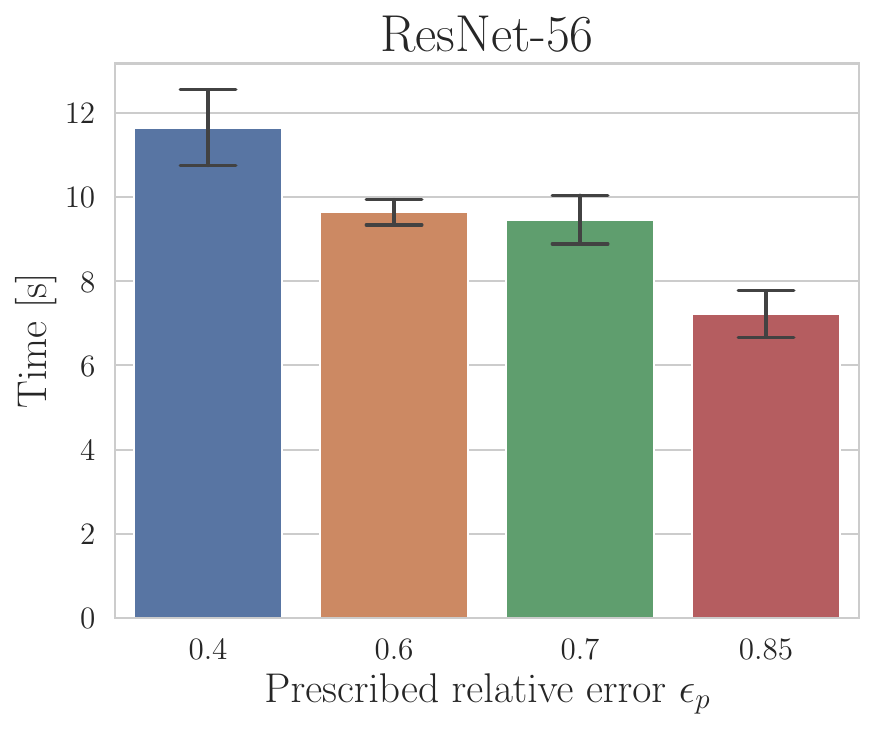}
    \end{subfigure}
    
    \begin{subfigure}[b]{\linewidth}
        \centering
        \includegraphics[scale=0.36]{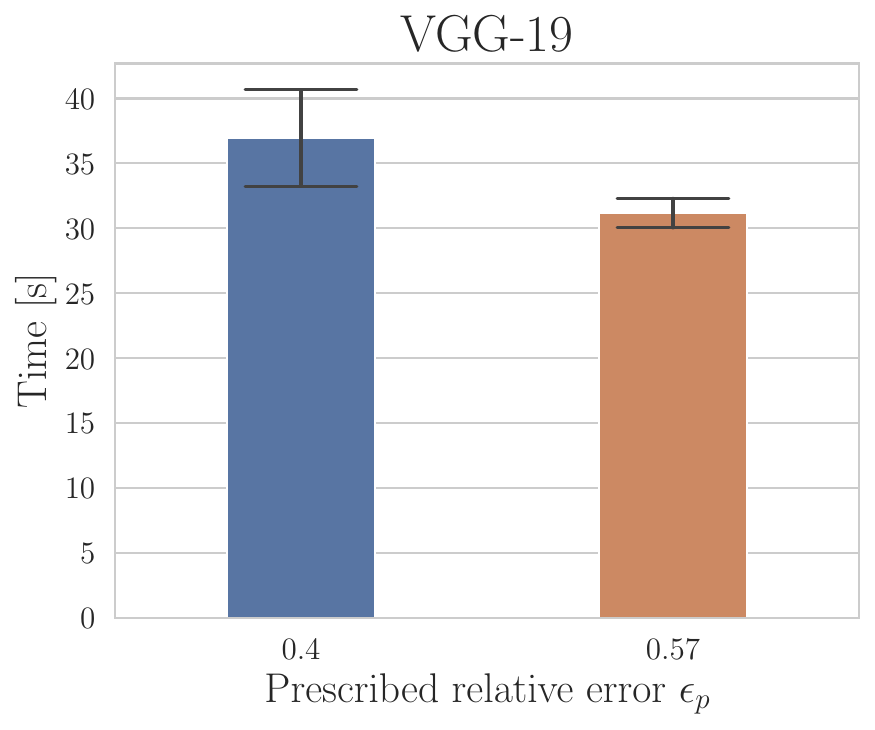}
        \hfill
        \includegraphics[scale=0.36]{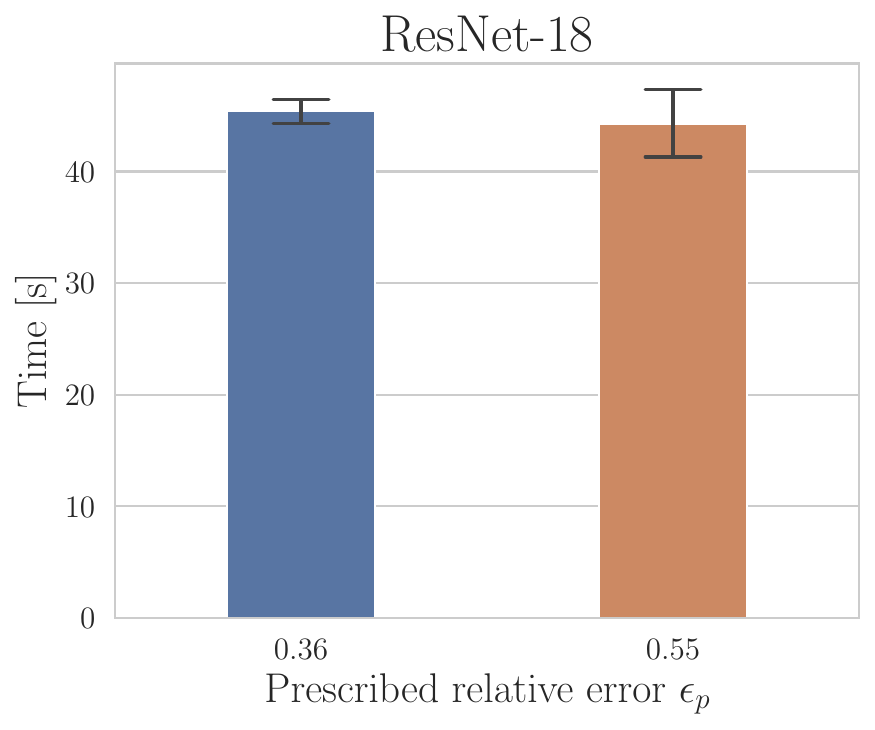}
        \hfill
        \includegraphics[scale=0.36]{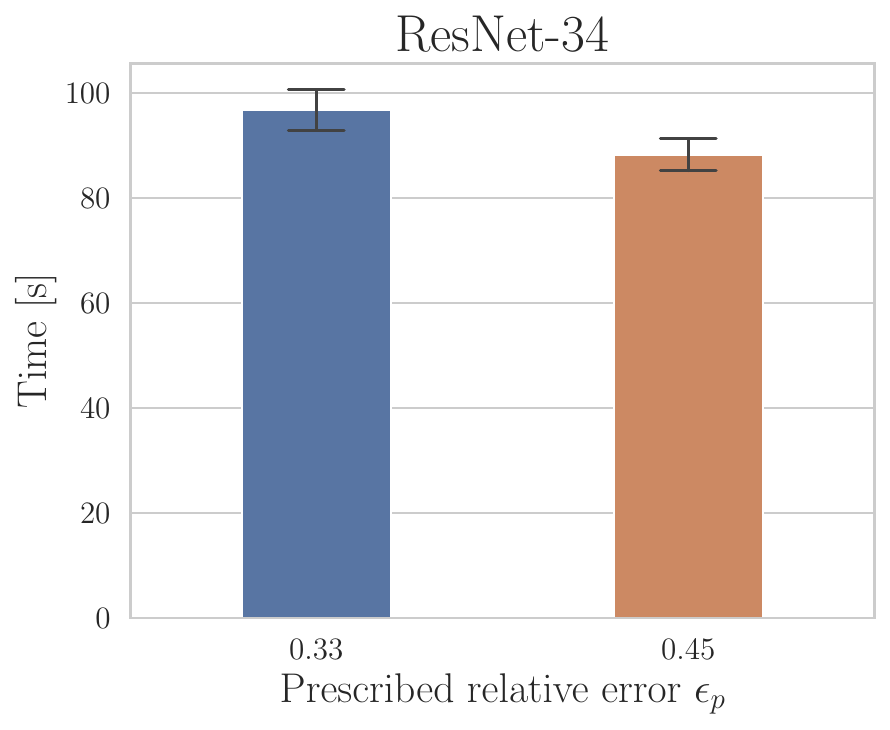}
    \end{subfigure}
    
    \caption{Compression times of evaluated CNNs averaged over 10 times.}
    \label{fig:compression_times}
\end{figure*}

\subsection{Discussion}
\label{sec:discussion}

The proposed method is characterized not only by strong parameter compression, but also by strong FLOPS compression, while preserving good classification quality. This is not the case in the previously proposed tensorized versions of TR, where high values of PCR are obtained at the cost of poorer network quality. For tensorized models, such as TR and PSTRN-S, the total number of FLOPS increases with respect to the baseline neural network. Furthermore, the ranks in other TR-based networks were basically determined manually (\cite{wang2018wide}) and were the same in all layers of the network. \cite{cheng2020novel} used the reinforcement learning (RL) to find the ranks, but only with respect to a layer and not to each mode. Moreover, the RL procedure is quite time-consuming. In the TR approach with the genetic algorithm,  \cite{li2021heuristic} explored the idea of finding TR ranks within each layer, but only for very small networks on the MNIST dataset. Additionally, the rank searching with genetic algorithms (\cite{li2021heuristic}) costs 2.5/3.2 GPUs days on Nvidia Tesla V100 for ResNet-20/32, which is quite expensive.

The proposed RSDTR is a TR approach that uses the TR decomposition algorithm in the CNN compression procedure. Therefore, compressed networks can be fine-tuned from decomposed factors instead of training them from scratch. To reduce storage complexity, the exhaustive search procedure is used to find an optimal scenario for the permutation and rank divisor in each convolutional layer.  
As a result, the best low-storage TR representation is selected at a given prescribed relative error. In the proposed approach, the ranks are calculated instantly without using additional time-consuming searching strategies. The time required for the compression of networks is shown in Figure \ref{fig:compression_times}. As can be noted from this figure, compression time for the evaluated networks is no larger than 97 s on the Intel Core i9-12900H CPU, which is marginal compared to fine-tuning, which can take few hours on multiple Nvidia V100 GPUs. Furthermore, the upper bounds of storage complexity for all possible circular-mode permutations are also computed analytically, and the most optimal case is suggested. The theoretical considerations are confirmed by the statistics of the exhaustive search procedure, which is demonstrated in the supplementary material. 

\section{Conclusions}
\label{sec:conclusions}
In this study, we proposed a novel low-rank approach to compress CNNs. The proposed RSDTR method for CNNs compression results in large parameter and FLOPS compression ratios, while preserving a good classification accuracy. It was evaluated using four CNNs of various sizes on the CIFAR-10 and ImageNet datasets. The experiments clearly show the efficiency of RSDTR compared to other state-of-the-art CNNs compression approaches, including tensorized TR approaches.

This study provides the foundation for extending the proposed method to compress higher-order CNNs (\cite{kossaifi2020factorized}), where convolutional kernels are represented as $n$-th-order tensors. Another line of research is to go beyond standard tensor factorization methods by searching tensor network structures (\cite{li2022permutation,li2023alternating,zheng2023svdinstn}) and finding an optimal one for each convolutional kernel. Furthermore, future studies should investigate the combination of the proposed approach with other compression methods, such as pruning. 

\section*{Acknowledgment}
This work is supported by National Center of Science under Preludium 22 project No. UMO-2023/49/N/ST6/02697 "Acceleration and compression of deep neural networks using low-rank approximation methods"

\section*{CRediT authorship contribution statement}
\textbf{Mateusz Gabor}: Conceptualization, Formal Analysis, Project Administration, Funding Acquisition, Investigation, Methodology, Software, Validation, Visualization, Writing – Original Draft Preparation, Writing – Review \& Editing.
 \textbf{Rafal Zdunek}: Conceptualization, Formal Analysis, Project Administration, Methodology, Resources, Supervision, Visualization, Writing – Original Draft Preparation, Writing – Review \& Editing.
\appendix

\section{Convolutions for circular shifts}
\label{appendix_convolutions}

The mapping in (\ref{conv_kr_tr}) holds for each circular permutation of the kernel weight tensor but the order (indices) and the sizes of core tensors change according to the following rules:
\begin{itemize}
\item $\mathcal{W}^{\tau_1} = \mathcal{W} \in \Real^{C \times D \times D \times T}$: $\mathcal{G}^{(1)} \rightarrow \mathcal{G}^{(4)} \in \Real^{R_4 \times T \times R_1}$,  $\mathcal{G}^{(2)} \rightarrow \mathcal{G}^{(1)} \in \Real^{R_1 \times C \times R_2}$, $\mathcal{G}^{(3)} \rightarrow \mathcal{G}^{(2)} \in \Real^{R_2 \times D \times R_3}$, and $\mathcal{G}^{(4)} \rightarrow \mathcal{G}^{(3)} \in \Real^{R_3 \times D \times R_4}$,
\item $\mathcal{W}^{\tau_2} = \mathcal{W} \in \Real^{D \times D \times T \times C}$: $\mathcal{G}^{(1)} \rightarrow \mathcal{G}^{(3)} \in \Real^{R_3 \times T \times R_4}$,  $\mathcal{G}^{(2)} \rightarrow \mathcal{G}^{(4)} \in \Real^{R_4 \times C \times R_1}$, $\mathcal{G}^{(3)} \rightarrow \mathcal{G}^{(1)} \in \Real^{R_1 \times D \times R_2}$, and $\mathcal{G}^{(4)} \rightarrow \mathcal{G}^{(2)} \in \Real^{R_2 \times D \times R_3}$,
\item $\mathcal{W}^{\tau_3} = \mathcal{W} \in \Real^{D \times T \times C \times D}$: $\mathcal{G}^{(1)} \rightarrow \mathcal{G}^{(2)} \in \Real^{R_2 \times T \times R_3}$,  $\mathcal{G}^{(2)} \rightarrow \mathcal{G}^{(3)} \in \Real^{R_3 \times C \times R_4}$, $\mathcal{G}^{(3)} \rightarrow \mathcal{G}^{(4)} \in \Real^{R_4 \times D \times R_1}$, and $\mathcal{G}^{(4)} \rightarrow \mathcal{G}^{(1)} \in \Real^{R_1 \times D \times R_2}$.
\end{itemize}
The change of order means that core tensor $\mathcal{G}^{(1)}$ in (\ref{conv_kr_tr}) for circular permutation $\tau_1$ needs to be replaced with $\mathcal{G}^{(4)}$, then,  $\mathcal{G}^{(2)}$ with $\mathcal{G}^{(1)}$, etc. Note that only the sizes of the core tensors change with circular permutations, but the ordering of mathematical operations in the respective mappings remains unchanged, which does not affect the pipeline structure in Fig. \ref{fig:dtr}.

\section{Tensorized TR}
\label{appendix_tensorized_TR}
\subsection{Model}
\label{appendix_tr_models}
\begin{figure}[!ht]
    \centering
    \includegraphics[scale=1]{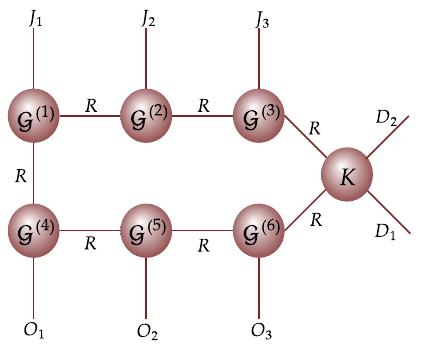}
    \caption{Tensorized TR convolutional layer.}
    \label{fig:tensorized_tr}
\end{figure}

The tensorized version of TR (\cite{wang2018wide}), which replaces the original convolution layer with a tensor network, is shown in Figure \ref{fig:tensorized_tr}.
This network consists of the following tensors: $\mathcal{G}^{(1)} \in \Real^{R \times J_1 \times R}$, $\mathcal{G}^{(2)} \in \Real^{R \times J_2 \times R}$, $\mathcal{G}^{(3)} \in \Real^{R \times J_3 \times R}$,
$\mathcal{G}^{(4)} \in \Real^{R \times O_1 \times R}$,
$\mathcal{G}^{(5)} \in \Real^{R \times O_2 \times R}$,
$\mathcal{G}^{(6)} \in \Real^{R \times O_3 \times R}$,
$\mathcal{K} \in \Real^{R \times R \times D \times D}$. 

Taking into account the input tensor $\mathcal{X} \in \Real^{I_1 \times I_2 \times C}$, then the convolution of tensorized TR format in Figure \ref{fig:tensorized_tr} is as follows. At the beginning, the input tensor is reshaped to the form $\tilde{\mathcal{X}} \in \Real^{I_1 \times I_2 \times J_1 \times J_2 \times J_3}$, where $C = J_1J_2J_3$. Then the following tensor is created:
\begin{equation}\label{contr1}
    \mathcal{Z} = \mathcal{G}^{(1)} \times_3^1 \mathcal{G}^{(2)} \times_4^1 \mathcal{G}^{(3)} \in \Real^{R\times J_1 \times J_2 \times J_3 \times R}
\end{equation}
This tensor is contracted with tensor $\tilde{\mathcal{X}}$ as follows:
\begin{equation}\label{contr2}
    \mathcal{V} = \tilde{\mathcal{X}}\times_{3,4,5}^{2,3,4}\mathcal{Z} \in \Real^{I_1 \times I_2 \times R \times R}.
\end{equation}
In the next step, 2D convolution with tensor $\mathcal{K}$ is performed:
\begin{equation}\label{trconv2d}
    \mathcal{V}^{(V,H)} = \mathcal{V} \times_4^1 \star_{1,2}^{3,4} \mathcal{K} \in \Real^{\tilde{I_1} \times \tilde{I_2} \times R \times R}.
\end{equation}
The tensors related to the output channels are contracted according to the following model:
\begin{equation}\label{contr3}
    \mathcal{S} = \mathcal{G}^{(4)} \times_3^1 \mathcal{G}^{(5)} \times_4^1 \mathcal{G}^{(6)} \in \Real^{R \times O_1 \times O_2 \times O_3 \times R}.
\end{equation}
Next, the following contraction is performed:
\begin{equation}\label{contr4}
    \mathcal{Y} = \mathcal{V}^{(V,H)} \times_{3,4}^{1,5}  \mathcal{S} \in \Real^{\tilde{I_1} \times \tilde{I_2} \times O_1 \times O_2 \times O_3 }.
\end{equation}
In the end, tensor $\mathcal{Y}$ is reshaped to the following form $\tilde{\mathcal{Y}} \in \Real^{\tilde{I_1} \times \tilde{I_2} \times T}$, where $T = O_1 O_2 O_3$.

\subsection{FLOPS complexity} \label{flops_tensorized}
Tensor $\mathcal{Z}$ in (\ref{contr1}) is created based on computing two contractions, which takes $\mathcal{O}(R^3J_1J_2 + R^3J_1J_2J_3)$ FLOPS. Multi-mode contraction in (\ref{contr2}) takes $\mathcal{O}(R^2J_1J_2J_3I_1I_2)$ FLOPS. The 2D convolution in (\ref{trconv2d}) is estimated as $\mathcal{O}(R^2D_1D_2I_1I_2)$ FLOPS. Tensor $\mathcal{S}$ in (\ref{contr3}) is built on performing two contractions, which costs $\mathcal{O}(R^3O_1O_2 + R^3O_1O_2O_3)$ FLOPS. The last contraction in (\ref{contr4}) takes $\mathcal{O}(R^2O_1O_2O_3\tilde{I_1}\tilde{I_2})$ FLOPS. Setting $C=J_1J_2J_3$ and $T = O_1O_2O_3$, the total number of FLOPS for the tensorized TR convolution is $\mathcal{O}(R^3(J_1J_2+C+T+O_1O_2) + R^2(CI_1I_2 + D_1D_2I_1I_2 + T\tilde{I_1}\tilde{I_2}))$.

\section{Storage complexity analysis}
\label{appendix:storage_complex}
Let $D = D_1 = D_2 \geq 3$, $D << \min \{T, C \}$, and $T \geq C$, which reflects real scenarios in CNNs. The upper bounds for storage complexities are computed, assuming full-rank $n$-unfoldings in TR-SVD (\cite{zhao2016tensor}). Let us consider all circular mode-permutation cases for weight tensor $\mathcal{W}$, which are illustrated in Fig. \ref{fig:TR_permutations}: 
\begin{itemize}
\item $\mathcal{W}^{\tau_0} = \mathcal{W} \in \Real^{T \times C \times D \times D}$: The storage complexity for the TR format of $\mathcal{W}^{\tau_0}$ is easy to compute if we know all TR ranks, and in this case it is given by 
\be \label{eq_storage_complex} \mathcal{O}_{{\tt TR}(\mathcal{W}^{\tau_0})} & = & R_1R_2T + R_2R_3C \nonumber \\ 
& + & R_3R_4D + R_4R_1D. \ee The TR ranks in TR-SVD can be computed based on the sequential SVD routine given truncation thresholds $\{ \delta_k\}$. However, the upper bounds for the storage complexity can be estimated without running TR-SVD, i.e. by roughly estimating maximal TR ranks, assuming $\delta_k = 0$ (usually for $k > 1$) and considering the sizes of decomposed matrices.  
Given $\mathcal{W}$ and $\delta_1$, TR-SVD (\cite{zhao2016tensor}) initializes the computations from 1-unfolding matrix $\bW_{<1>} \in \Real^{T \times CD^2}$. Thus, $R_1R_2 \leq \min \{T, CD^2 \}$. Solving the problem $\min_{R_1,R_2}||R_1 - R_2||$, s.t. $R_1R_2 = {\rm rank}_{\delta_1}(\bW_{<1>})$, we have $R_2 = \frac{{\rm rank}_{\delta_1}(\bW_{<1>})}{R_1}$. Assuming $\delta_1 = 0$, which corresponds to the non-truncated case (assuming full-rank unfoldings), we have $R_2 = \frac{T}{R_1}$, if $CD^2 \geq T$. This assumption is satisfied in practice because usually $C \geq \frac{T}{4}$. In the next step of the sequential SVD routine (computation of $\mathcal{G}^{(2)}$), $R_3$ needs to be estimated. However, note that $R_3 \leq \min \{ D^2 R_1, R_2 C \}$. The condition $R_2C \geq D^2R_1$ is satisfied for $R_1 \in \left [1,\frac{\sqrt{TC}}{D} \right]$, and in this case the upper bound for $R_3$ is $D^2R_1$. Otherwise, $R_3 = \frac{TC}{R_1}$. In the last step (computation of $\mathcal{G}^{(3)}$ and $\mathcal{G}^{(4)}$), $R_4 \leq \min \{ R_3D, R_1D\}$. Hence, $R_4 = R_1D$ for $R_1 \in \left [1, \sqrt{TC} \right ]$, or $R_4 = \frac{TCD}{R_1}$ for $\sqrt{TC} < R_1 \leq T$. After inserting the above upper bounds for $R_2$, $R_3$, and $R_4$ to (\ref{eq_storage_complex}), and perforing straightforward computations, the storage complexity of the TR format of $\mathcal{W}^{\tau_0}$ can be expressed by $\mathcal{O}_{{\tt TR}(\mathcal{W})}^{(1)} = T^2 + TD^2C + (D^4 + D^2)R_1^2$ for $R_1 \in \left [1,\frac{\sqrt{TC}}{D} \right ]$. For $R_1 \in \left [\frac{\sqrt{TC}}{D}, \sqrt{TC} \right )$, $\mathcal{O}_{{\tt TR}(\mathcal{W})}^{(2)} = T^2 + \left (\frac{TC}{R_1} \right )^2  + TCD^2 + (DR_1)^2$, and $\mathcal{O}_{{\tt TR}(\mathcal{W})}^{(3)} = T^2 + \left (\frac{TC}{R_1} \right )^2  + \left (\frac{TCD}{R_1} \right )^2 + TCD^2$ for $R_1 \in \left [\sqrt{TC}, T \right]$. 
\item $\mathcal{W}^{\tau_1} \in \Real^{C \times D \times D \times T}$: For this case, $R_1R_2 \leq \min \{ C, D^2T\}$, $R_3 \leq R_2D$, and $R_4 \leq \min \{ R_3 D, T R_1\}$. Following a similar full-rank assumption as for $\mathcal{W}^{\tau_0}$, the upper bound for a storage complexity of $\mathcal{W}^{\tau_1}$ in the TR format is given by: $\mathcal{O}_{{\tt TR}(\mathcal{W}^{\tau_1})}^{(1)} = C^2 + \left ( \frac{CD}{R_1}\right )^2 + CTD^2 + (TR_1)^2$ for $R_1 \in \left [1, D\sqrt{\frac{C}{T}} \right ]$, and  $\mathcal{O}_{{\tt TR}(\mathcal{W}^{\tau_1})}^{(2)} = C^2 + \left ( \frac{CD}{R_1}\right )^2 + \left ( \frac{CD^2}{R_1}\right )^2 + CTD^2$ for $R_1 \in \left (D\sqrt{\frac{C}{T}}, C \right ]$.
\item $\mathcal{W}^{\tau_2} \in \Real^{D \times D \times T \times C}$: Since $R_1$ is a divisor of ${\rm rank}_{\delta}(\bW_{<1>}^{\tau_2})$, $TC >> 1$, $R_2 = \frac{D}{R_1}$, and $D$ is usually an odd number, typically not exceeding 5, therefore, the only choice for $R_1$ in this case is $R_1 = 1$ or $R_1 = D$.  For $R_1 = 1$, we have $R_2 = D$, $R_3 = D^2$, $R_4 = C$, and $\mathcal{O}_{{\tt TR}(\mathcal{W}^{\tau_2})}^{(1)} = D^2 + D^4 + CTD^2 + C^2$. For $R_1 = D$, $R_2 = 1$, $R_3 = D$, $R_4 = DC$, and $\mathcal{O}_{{\tt TR}(\mathcal{W}^{\tau_2})}^{(2)} = 2D^2 + CTD^2 + (DC)^2$. 
\item $\mathcal{W}^{\tau_3} \in \Real^{D \times T \times C \times D}$: Similarly to $\mathcal{W}^{\tau_2}$, for $R_1 = 1$, $R_2 = D$, and $R_3 = T$, we have $\mathcal{O}_{{\tt TR}(\mathcal{W}^{\tau_3})}^{(1)} = 2D^2 + CTD^2 + (DC)^2$; and $\mathcal{O}_{{\tt TR}(\mathcal{W}^{\tau_3})}^{(2)} = D^2 + D^4 + CTD^2 + T^2$ for $R_1 = D$. 
\end{itemize}

\begin{figure}[!ht]
    \centering
    \includegraphics[scale=0.22]{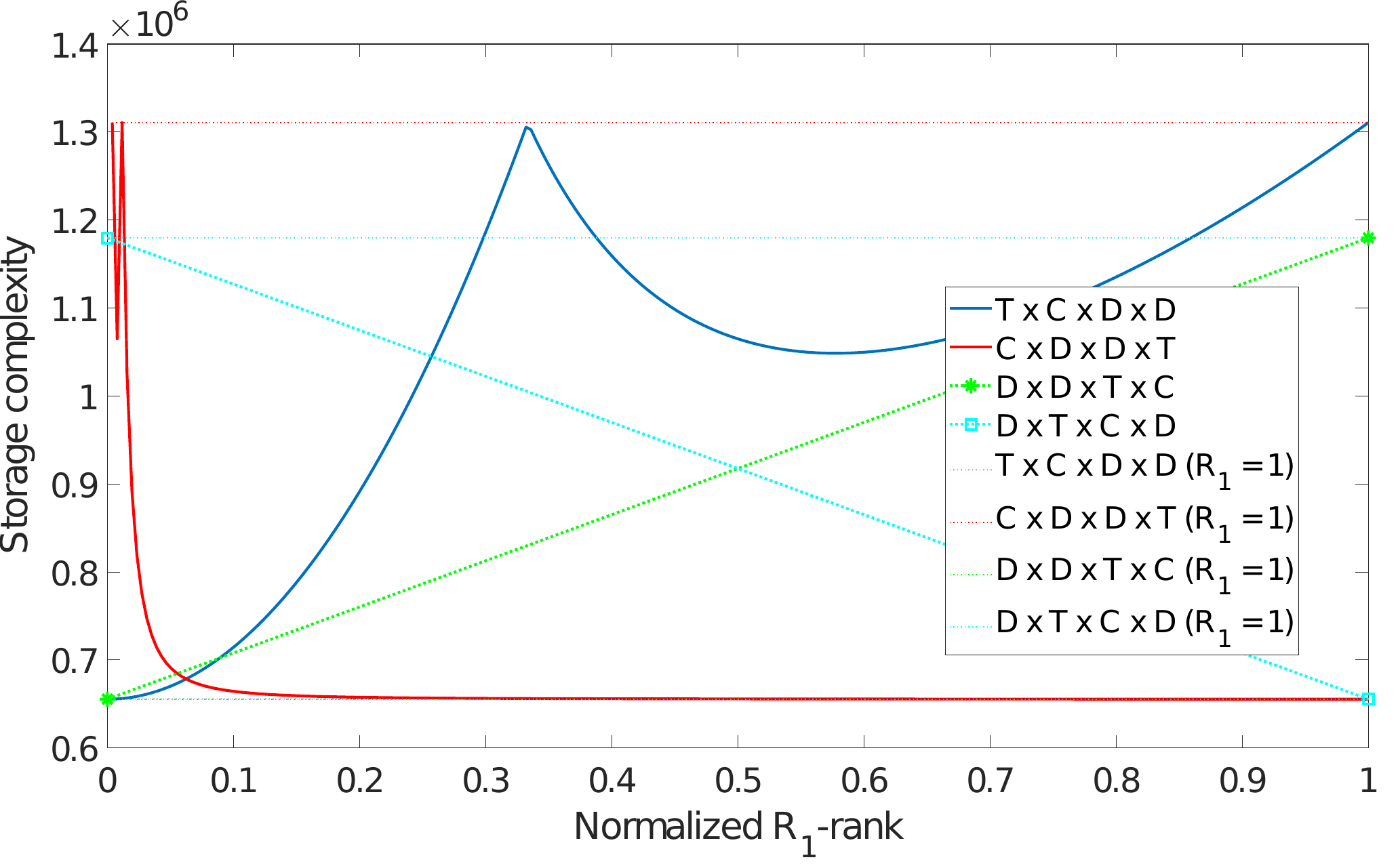}
    \caption{Storage complexity versus normalized $R_1$-rank for all circular mode-permuations. Parameters: $T = C = 256$, $D = 3$.} 
    \label{fig:sc_256}
\end{figure}

\begin{figure}[!ht]
    \centering
    \includegraphics[scale=0.22]{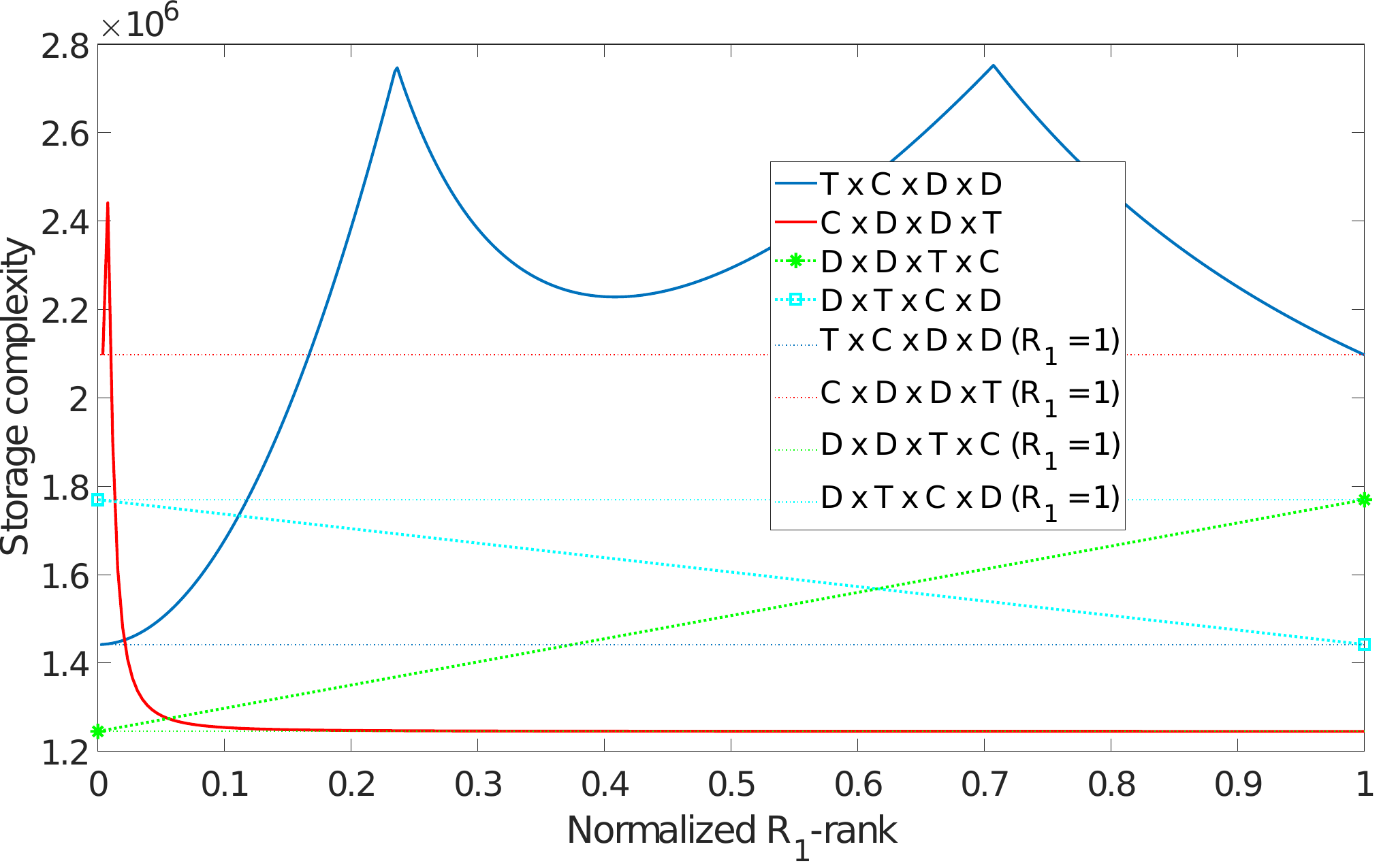}
    \caption{Storage complexity versus normalized $R_1$-rank for all circular mode-permuations. Parameters: $T = 512$, $C = 256$, $D = 3$.} 
    \label{fig:sc}
\end{figure}

All storage complexities for the above-discussed cases, where $T = C = 256$, and $D = 3$, are plotted in Fig. \ref{fig:sc_256} versus the normalized $R_1$-rank, which means $\tilde{R_1} \leftarrow \frac{R_1}{\max(R_1)}$ for each permutation case. When $T = 512$, $C = 256$, and $D = 3$, the storage complexities are depicted in Fig. \ref{fig:sc}. Note that in both cases $\mathcal{O}_{{\tt TR}(\mathcal{W}^{\tau_0})}^{(3)}(R_1 = T) = \mathcal{O}_{{\tt TR}(\mathcal{W}^{\tau_1})}^{(1)}(R_1 = 1)$, $\mathcal{O}_{{\tt TR}(\mathcal{W}^{\tau_1})}^{(2)}(R_1 = C) = \mathcal{O}_{{\tt TR}(\mathcal{W}^{\tau_2})}^{(1)}(R_1 = 1)$, $\mathcal{O}_{{\tt TR}(\mathcal{W}^{\tau_2})}^{(2)}(R_1 = D) = \mathcal{O}_{{\tt TR}(\mathcal{W}^{\tau_3})}^{(1)}(R_1 = 1)$, and $\mathcal{O}_{{\tt TR}(\mathcal{W}^{\tau_3})}^{(2)}(R_1 = D) = \mathcal{O}_{{\tt TR}(\mathcal{W}^{\tau_0})}^{(1)}(R_1 = 1)$. These equivalences are marked in Figs. \ref{fig:sc_256} and \ref{fig:sc} with dot lines. 
\begin{figure*}[!ht]
    \begin{subfigure}[b]{\linewidth}
        \centering
        \includegraphics[scale=0.365]{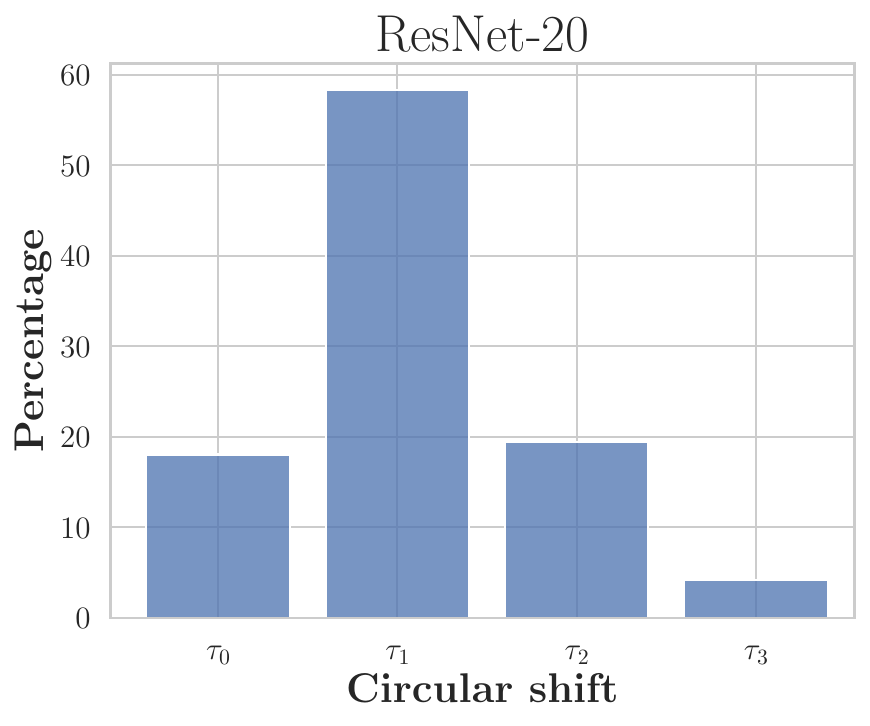}
        \hfill
        \includegraphics[scale=0.365]{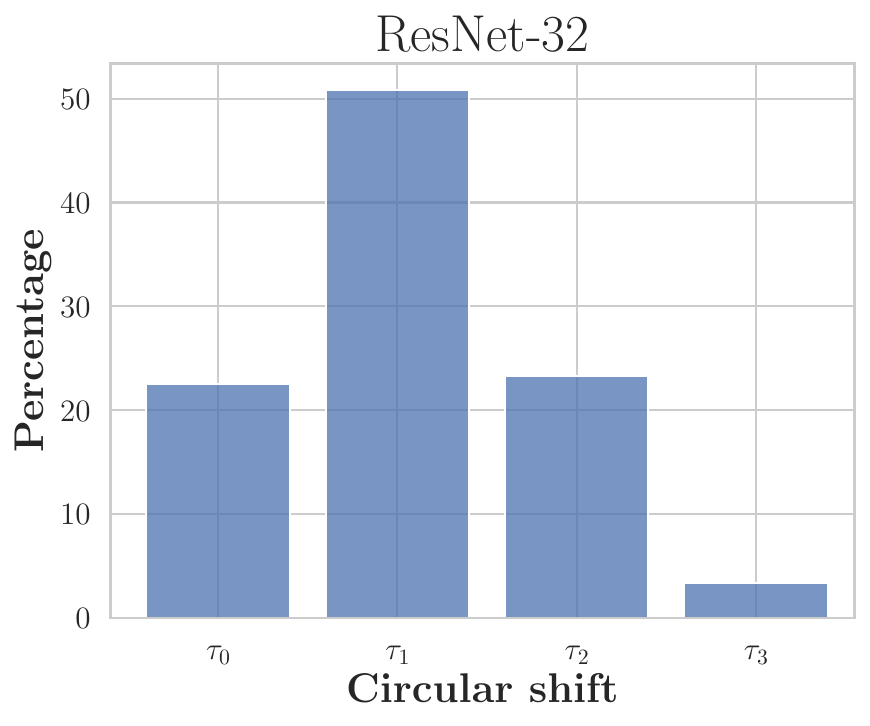}
        \hfill
        \includegraphics[scale=0.365]{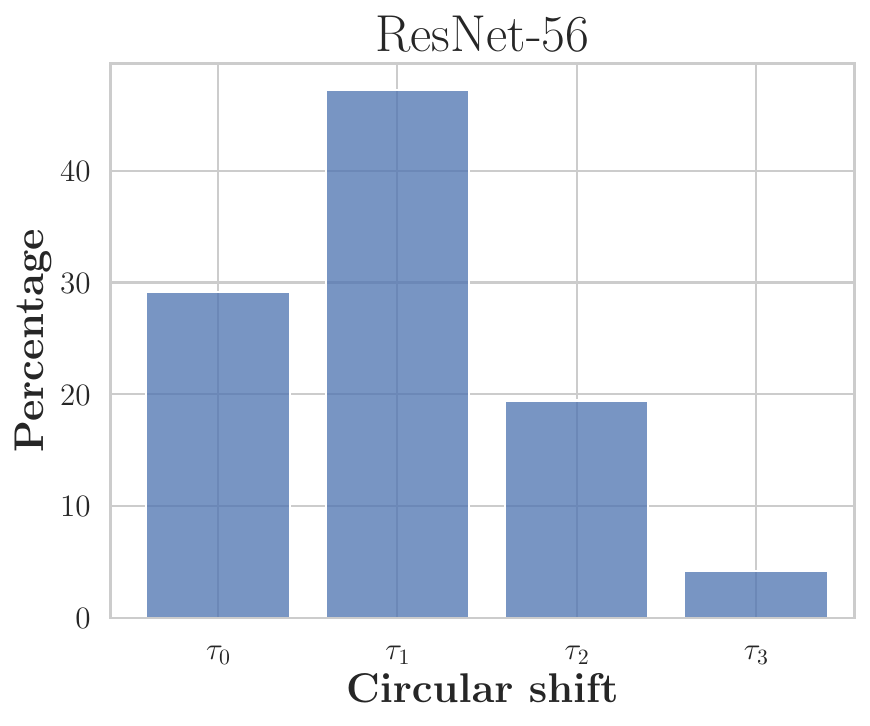}
    \end{subfigure}
    
    \begin{subfigure}[b]{\linewidth}
        \centering
        \includegraphics[scale=0.365]{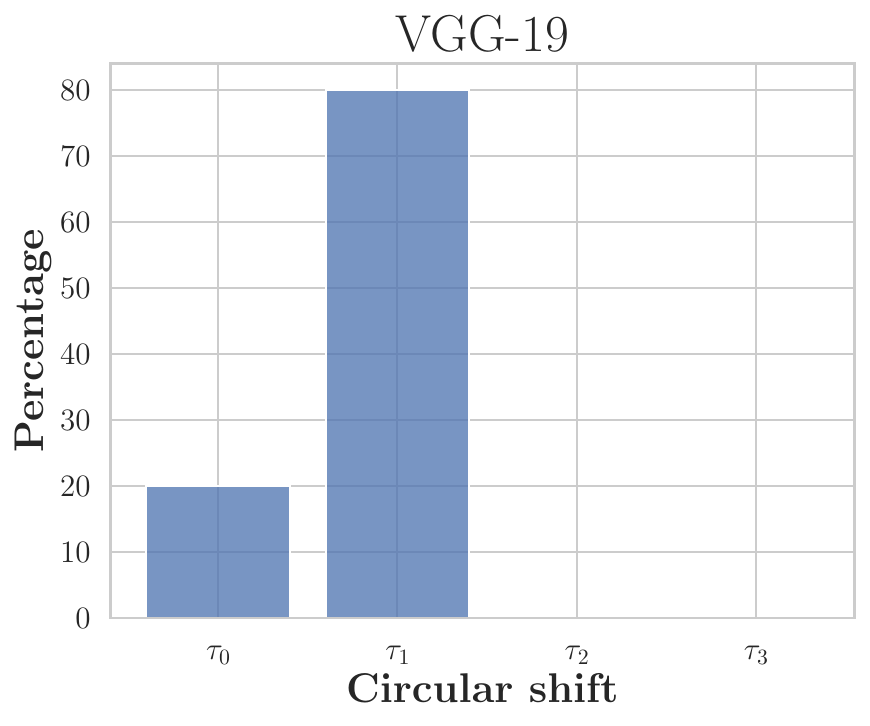}
        \hfill
        \includegraphics[scale=0.365]{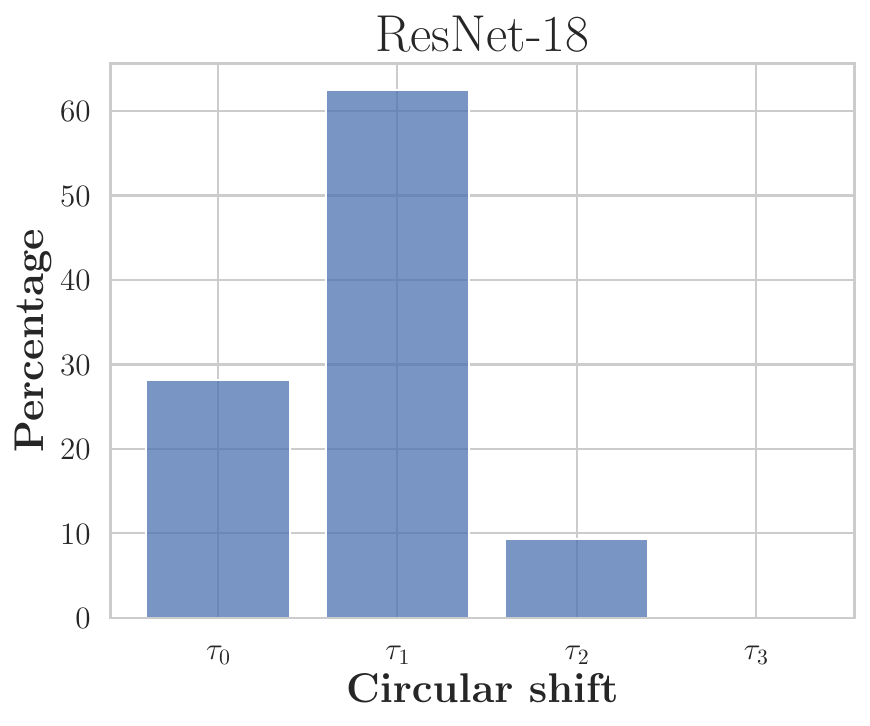}
        \hfill
        \includegraphics[scale=0.365]{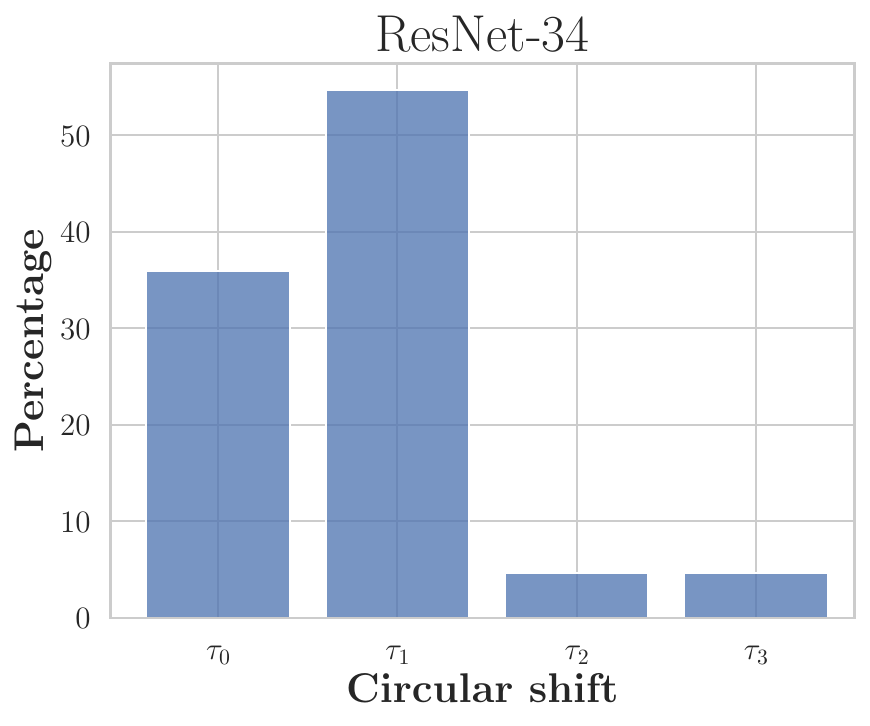}
    \end{subfigure}
    
    \caption{Relative frequencies of circular mode permutations selected by the RSDTR algorithm in all test cases.}
    \label{fig:permuations}
\end{figure*}
\begin{remark}
\label{r1}
Considering the above calculated upper bounds for storage complexities, we can conclude that the lowest storage complexity depends on the ratio of $\frac{T}{C}$ and can be obtained for a few configurations, i.e. circular mode-permutations and $R_1$-rank. If $T = C$ (Fig. \ref{fig:sc_256}), it occurs for: (a) $\mathcal{W}^{\tau_0}$ and $R_1 = 1$, (b) $\mathcal{W}^{\tau_1}$ and $R_1 = C$, (c) $\mathcal{W}^{\tau_2}$ and $R_1 = 1$, and (d) $\mathcal{W}^{\tau_3}$ and $R_1 = D$. Each case corresponds to the TT format because case (b) mathematically boils down to case (c), and case (d) to case (a). Despite the advantage in storage complexity, the TT ranks reflect a fixed pattern (lower for the outer cores and higher for the middle cores), and hence, the TT representations are not optimal to capture the most informative features. For $\mathcal{W}^{\tau_1}$, the storage complexity function (red line) is nearly flat in a wide range of $R_1$, thus the circular permutation $C \times D \times D \times T$ seems to be the best choice for $R_1 \in \left (\frac{CD^2}{\sqrt{T^2 + D^4 - C^2}}, C \right ]$. In fact, this pattern most often emerges empirically in the experiments carried out in this study (see Fig. \ref{fig:permuations}). When $T = 2C$ (Fig. \ref{fig:sc}), the lowest storage complexity occurs only for cases (b) and (c), which are mathematically equivalent. Thus, the same conclusions can be drawn for this case. 
\end{remark}

\bibliographystyle{elsarticle-harv} 
\bibliography{references}

\end{document}